\documentclass[10pt,journal,compsoc]{IEEEtran}
\usepackage[T1]{fontenc}

\usepackage{times}
\usepackage{epsfig}
\usepackage{graphicx}
\usepackage{amsmath}
\usepackage{amssymb}
\usepackage{booktabs}
\usepackage{algorithmic}
\usepackage[linesnumbered,ruled]{algorithm2e}
\usepackage{subfigure}
\usepackage{bm}
\usepackage{multirow}
\usepackage{color}
\usepackage{subfigure}

\newcommand{\vect}[1]{\mathbf{#1}}
\newcommand{\gvect}[1]{\boldsymbol{#1}}
\newcommand{\sketch}{\textsc{s}}
\newcommand{\image}{\textsc{i}}
\newcommand{\joint}{\textsc{j}}

\def\eg{\textit{e.g.\ }}
\def\ie{\textit{i.e.\ }}

\def\etal{\textit{et al.\ }}

\newcommand{\tabincell}[2]{\begin{tabular}{@{}#1@{}}#2\end{tabular}}

\ifCLASSOPTIONcompsoc
  \usepackage[nocompress]{cite}
\else
  \usepackage{cite}
\fi

\ifCLASSINFOpdf
\else
\fi
\begin{document}
%
\title{\huge{Cross-Paced Representation Learning with Partial Curricula for Sketch-based Image Retrieval}}
%
%
%

\author{Dan Xu,~\IEEEmembership{Student Member,~IEEE}, Xavier Alameda-Pineda,~\IEEEmembership{Member,~IEEE}, Jingkuan Song, Elisa 
Ricci,~\IEEEmembership{Member,~IEEE}
        and~Nicu Sebe,~\IEEEmembership{Senior~Member,~IEEE}%
\IEEEcompsocitemizethanks{\IEEEcompsocthanksitem Dan Xu and Nicu Sebe are with the Department
of Information Engineering and Computer Science, University of Trento, Italy. (Email: \{dan.xu, 
niculae.sebe\}@unitn.it) \protect
\IEEEcompsocthanksitem Xavier Alameda-Pineda is with the Perception Team at INRIA, France. (Email: xavier.alameda-pineda@inria.fr)\protect
\IEEEcompsocthanksitem Jingkuan Song is with the School of Engineering and Applied Science, Columbia University, USA. (Email: jingkuan.song@gmail.com)\protect
\IEEEcompsocthanksitem Elisa Ricci is with Fondazione Bruno Kessler and University of Trento, Italy. (Email: eliricci@fbk.eu)}
\thanks{Manuscript received April 19, 2005; revised August 26, 2015.}}

%
%

\markboth{Journal of \LaTeX\ Class Files,~Vol.~14, No.~8, August~2015}%
{Shell \MakeLowercase{\textit{et al.}}: Bare Demo of IEEEtran.cls for IEEE Communications Society Journals}
%



\IEEEtitleabstractindextext{%
\begin{abstract}
In this paper we address the problem of learning robust cross-domain representations for sketch-based image retrieval 
(SBIR). While most SBIR approaches focus on extracting low- and mid-level descriptors for direct feature matching, 
recent works have shown the benefit of learning coupled feature representations to describe data from two related 
sources. However, cross-domain representation learning methods are typically cast into non-convex minimization 
problems that are difficult to optimize, leading to unsatisfactory performance. Inspired by self-paced learning,
a learning methodology designed to overcome convergence issues related to local optima by exploiting the samples in a 
meaningful order (\ie~easy to hard), we introduce the cross-paced partial curriculum learning (CPPCL) framework. 
Compared with existing self-paced learning methods which only consider a single modality and cannot deal with prior 
knowledge, CPPCL is specifically designed to assess the learning pace by jointly handling data from dual sources and 
modality-specific prior information provided in the form of partial curricula.
Additionally, thanks to the learned dictionaries, we demonstrate that the proposed CPPCL embeds robust coupled representations 
for SBIR.
Our approach is extensively evaluated on four publicly available datasets (\ie~CUFS, Flickr15K, QueenMary SBIR and TU-Berlin Extension datasets), showing superior performance over competing SBIR methods.
\end{abstract}

\begin{IEEEkeywords}
SBIR, Cross-domain Representation Learning, Self-paced Learning, Coupled Dictionary Learning.
\end{IEEEkeywords}}

\maketitle

\IEEEdisplaynontitleabstractindextext

%
\IEEEpeerreviewmaketitle

\section{Introduction}
\label{intro}
%
%
%
%
In the last few years, the developments in mobile device applications have increased the demand for powerful and 
efficient tools to query large-scale image databases. In particular, favored by the widespread diffusion of consumer 
touchscreen devices, sketch-based image retrieval (SBIR) has gained popularity. Most prior works on 
SBIR~\cite{hu2013performance,saavedra2014sketch,eitz2010evaluation,saavedrasketch,saavedra2010improved} focused on 
designing low- and mid-level features, and used the same type of descriptors for representing both sketches and image 
edge maps, allowing a direct matching between the two modalities. However, these methods implicitly assume that
the statistical distributions of image edges and sketches are similar.
Unfortunately, this assumption does not hold in many applications. Therefore, more recent studies proposed to use 
different feature descriptors to better represent the different modalities and learned a shared feature space using 
cross-domain representation learning methods. In particular, recent approaches based on dictionary learning 
(DL)~\cite{lee2006efficient,yang2010image,wang2012semi,huang2013coupled} or deep networks~\cite{feng2014cross, wang2017adversarial, xu2017learning, xu2017multi} have 
been proven especially successful for learning coupled representations from cross-modal data. However, these methods are 
usually based on non-convex optimization problems and can get easily stuck into local optima, with an adverse impact on 
the representational power and generalization capabilities of the learned descriptors.

\begin{figure}[t]
\centering
\includegraphics[width=0.435\textwidth]{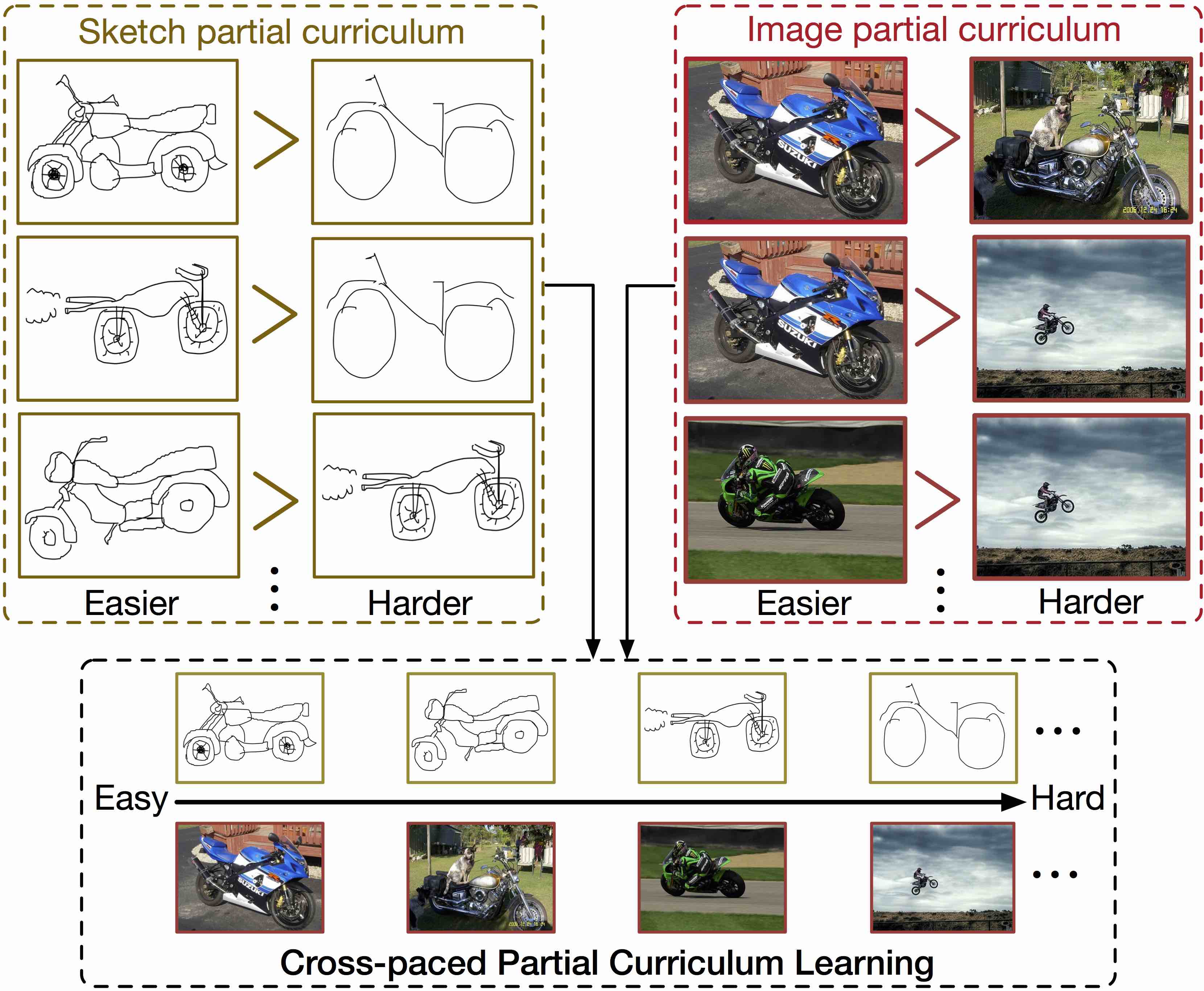} 
\caption{In real SBIR scenarios, both sketches and images show a wide range of visual complexity. Defining 
\textit{a-priori} a full learning order (\ie for all samples) based on the difficulty of the sketches/images is 
extremely challenging. Cross-paced partial curriculum learning combines the flexibility of partial modality-specific 
curricula with the power of self-paced learning strategies to automatically construct a full sample learning order that 
evolves over time until all training samples are used for learning.}
\label{fig:intro}
\vspace{-13pt}
\end{figure}%

Recent research efforts to overcome the problems associated to local optima resulted in two orthogonal trends: 
self-paced learning (SPL)~\cite{kumar2010self} and curriculum learning (CL)~\cite{bengio2009curriculum}. The common 
denominator of both SPL and CL is to build a learning model with the help of a sample order reflecting the inherent 
data complexity. The rationale is that, when this order is appropriately chosen, we increase the chances of avoiding 
local minima. SPL and CL have been successfully applied to several computer vision tasks, such as object 
tracking~\cite{supancic2013self} and visual category discovery~\cite{lee2011learning}. Even if both strategies share a 
common denominator, they are quite different in spirit. Indeed, while in CL the learning order is pre-determined by an expert or 
according to other prior knowledge (\eg extracted from the data), in SPL the algorithm automatically assesses the 
learning order usually based on the feedback of the learned model. Recently, Jiang \etal~\cite{jiang2015self} 
demonstrated that further advantages in terms of performance can be obtained by combining CL and SPL.

The particular case of SBIR is of special interest regarding CL, SPL and possible combinations. Indeed, as shown in 
Fig.~\ref{fig:intro}, the visual complexity of sketches and images greatly varies, and methods attempting to exploit 
these variations would a priori have more chances to successfully learn efficient and robust cross-domain 
representations. Specifically, natural images are characterized by cluttered background and objects-of-interest 
captured at different scales or various poses. Similarly, sketches drawn by expert/non-expert show remarkable 
variations. Therefore, our aim is to turn what could be seen as an adversity, into an exploitable feature inherent to 
the data. However, there are two major problems which hinder the direct application of existing SPL and CL methods into 
cross-domain representation learning models for SBIR. Firstly, the SBIR task involves data from two different 
modalities, while most of the previous SPL and CL approaches are fundamentally designed to model data from a single 
modality. Secondly, CL methods assume the existence of a full curriculum (\ie a complete order of all samples). This 
limits the applicability of CL methods to small/medium-scale problems, since the curriculum is usually designed by 
humans and assessing the easiness order of all samples (images and sketches) would be a chimerically 
resource-consuming task.

To address these problems, we design a novel cross-modality representation learning paradigm and apply it to the SBIR 
task. In details, we propose a novel self-paced learning strategy able to handle cross-modal data and to incorporate 
incomplete prior knowledge (\ie~partial modality-specific curricula), and we name it Cross-Paced Partial Curriculum 
Learning (CPPCL). Furthermore, we embed this strategy into a coupled dictionary learning framework for computing robust 
cross-domain representations. Specifically, our method learns a pair of image- and sketch-specific dictionaries, 
together with the associated sparse codes, enforcing the similarity between the codes of corresponding sketches and 
images. The reconstruction loss with the learned dictionaries, the code correspondence and the partial modality-specific 
curricula jointly determine which samples to learn from. We extensively evaluate our cross-domain representation 
learning on four publicly available datasets (\ie \ CUFS, Flickr15K, QueenMary SBIR, TU-Berlin Extension), demonstrating the 
effectiveness of the proposed learning strategy and achieves superior performance over competing SBIR approaches. The 
main contributions of this paper are:
\begin{itemize}
 \item We introduce the cross-paced partial curriculum learning paradigm to effectively integrate the self-pacing 
philosophy with modality-specific partial curricula and investigate different self-paced regularizers.
\item We propose an instantiation of CPPCL within the framework of coupled dictionary learning to obtain robust 
cross-domain representations for SBIR and we develop an efficient algorithm to learn the modality-specific dictionaries 
and codes, while assessing the optimal learning order jointly from the partial curricula and the representation power 
of the model at the current iteration.
\item We carry out an extensive experimental evaluation and analysis of the whole cross-domain representation
learning framework, exhibiting its effectiveness for SBIR on four different publicly available datasets.
\end{itemize}

The paper extends our conference submission \cite{xu2016academic} by reformulating the proposed CPPCL considering 
different self-paced regularization terms (\eg adding Self-paced regularizer A in Section \ref{subsec:CPRL}) and developing the associated optimization algorithms (Section
\ref{sec:optim}). From the experiments perspective, we discuss the influence, similarities and differences when using 
the different regularizing schemes within the proposed cross-paced learning framework on two publicly available datasets. 
A more in-depth analysis is conducted to further show the effectiveness of the proposed approach, including some parameter sensitivity study 
and a convergence analysis of different models (Section \ref{experiments}). Moreover, the introduction and related works parts are reorganized and significantly extended.

The rest of the paper is organized as follows: we first review 
the related work in Section~\ref{sec:related}, and then elaborate the details of the proposed approach and associated 
optimization algorithms in Sections~\ref{sec:method} and~\ref{sec:optim} respectively. The experimental results are 
presented in Section~\ref{sec:exps} and we conclude the paper in Section~\ref{sec:conclusion}.

\section{Related Work}\label{sec:related}
\begin{figure*}[t]
\centering
\includegraphics[width=0.92\textwidth]{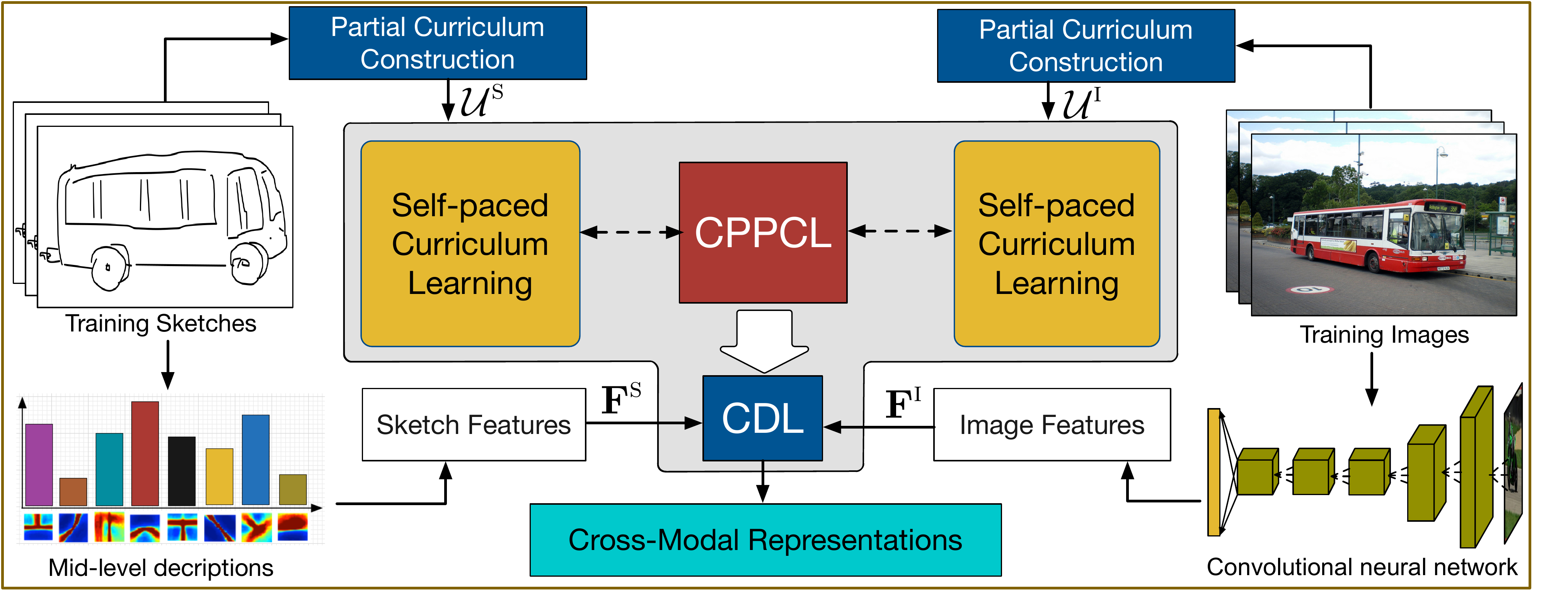} 
\caption{Overview of the proposed cross-modal representation learning method. Features extracted from sketches (\eg LKS descriptors) and images 
(\eg CNN-derived representations) are employed within a coupled dictionary learning (CDL) framework for computing
cross-modal representations for SBIR. Our CDL integrates a novel cross-paced partial 
curriculum learning paradigm which allows the learning algorithm to start with easy samples and gradually involve hard 
samples according to predetermined heuristics (\ie~modality-specific partial curricula).}
\label{framework}
\vspace{-13pt}
\end{figure*}

This section reviews related works in the areas of: (i) sketch-based image retrieval, 
(ii) self-paced and curriculum learning and (iii) cross-domain dictionary learning.

\subsection{Sketch-based Image Retrieval}
SBIR approaches are mostly based on matching feature descriptors of the query sketch with those of the edge maps 
associated to the images in the database. Early works on SBIR attempted to use existing low-level feature 
representations (\eg~describing color, texture, contour and shape) for both the sketch and the image modalities. Both 
global low-level descriptors (\eg~color histograms~\cite{kato1992sketch}, distribution of edge 
pixels~\cite{chalechale2005sketch}, elastic contours~\cite{bimbo1997visual}) and local ones (\eg~spark 
descriptors~\cite{eitz2011sketch}, SYM-FISH~\cite{cao2013sym}, SIFT~\cite{lowe1999object}, 
HOG~\cite{dalal2005histograms}) were investigated in the literature. Other works focused on developing specific 
descriptors for SBIR. For instance, Hu~\etal~\cite{hu2013performance} introduced the Gradient Field HOG (GF-HOG) 
descriptor, extending HOG to better represent sketches, and constructed a large dataset for evaluation: the Flickr15K. 
Saavendra~\etal~\cite{saavedra2010improved} also proposed a modified version of HOG, the histogram of edge 
local orientations (HELO), to tackle the problem of sparsity arising when HOG descriptors are 
applied to sketches. 

To represent sketches or image edges more robustly, most recent SBIR methods focused on constructing mid- or high-level feature descriptors. Several works considered the bag-of-words (BoW) technique to aggregate low-level features and generate 
mid-level representations~\cite{eitz2011sketch, sun2013indexing, lin20133d}. In addition to BoW-based methods, other 
approaches also focused on mid-level representations. For instance, in~\cite{saavedrasketch} an effective method 
to generate mid-level patterns, named learned keyshapes (LKS), was proposed for representing sketches. Yi~\etal\cite{li2014fine} built mid-level representations for both sketches and images by optimizing a deformable 
part-based model. Xiao~\etal\cite{xiao2015sketch} designed a shape feature descriptor especially useful for preserving 
the shape information of sketches. A perceptual grouping framework was introduced in~\cite{qi2015making} to organize 
image edges into a meaningful structure and was adopted for generating human-like sketches useful for SBIR. Yu~\etal\cite{yu2015sketch, yu2016sketch} 
proposed to adopt deep CNNs to learn high-level sketch representations. Similarly, Liu~\etal\cite{liu2017deep} explored deep 
representations within a binary coding framework for fast sketch based image retrieval.
 
In all these works the same low- and mid/high-level representations are used to describe both the sketch and image 
modalities, such as to facilitate direct feature matching. However, due to the difference in appearance between 
sketches and images, different features are more suitable to represent the two modalities. Following this idea, some 
works proposed learning a shared feature space for the two modalities~\cite{wang2013learning,xu2016instance}.
However, none of these works considered exploiting the visual complexity of samples to learn more effective cross-modal 
feature representations. 


\subsection{Self-paced and Curriculum Learning}
Inspired by the way the human brain explores the world, \ie starting from easy concepts first and gradually involving 
more complex notions, self-paced learning~\cite{kumar2010self} and curriculum learning~\cite{bengio2009curriculum} have 
been recently developed. The idea of SPL and CL is to learn models  in an incremental fashion from samples with 
variate difficulty presented in a meaningful order. Due to their generality, these techniques have been considered in a 
broad spectrum of learning tasks and models, including matrix factorization~\cite{zhao2015self,jiang2015self}, 
clustering~\cite{xumulti}, multi-task learning~\cite{pentina2014curriculum} and dictionary 
learning~\cite{tang2012self,xu2016instance}. They have also shown to be successful in many computer vision applications 
such as object tracking~\cite{supancic2013self}, media retrieval~\cite{jiang2014easy}, 
visual category discovery~\cite{lee2011learning} and event detection~\cite{jiang2014self}. 

Although self-paced learning and curriculum learning develop from the same rationale, they differ in the specific 
implementation schemes. In CL, the learning order (\ie~the curriculum) is pre-defined according to prior knowledge and 
fixed during the learning phase, while in SPL the curriculum is dynamically determined based on the feedback from the 
learner. 
Since the sample order in SPL is dynamically inferred, one challenging task is to design a meaningful strategy of 
assessing the difficulty of the training samples. Previous works have addressed this issue in different ways. The most 
common strategy is to measure the easiness of a sample by computing the associated loss~\cite{kumar2010self}. 
Alternatively, Jiang~\etal\cite{jiang2014easy} proposed to take into account the dissimilarity with respect to what has 
already been learned. To incorporate the benefits of both SPL and CL, a recent work~\cite{jiang2015self} proposed a self-paced 
curriculum learning framework in which the learning order is jointly determined by a predefined full-order curriculum 
and the learning feedback. However, none of these previous works focused on handling multi-modal data. Our approach not 
only extends the self-paced learning paradigm to cope with cross-domain data, but, more importantly, is naturally able 
to utilize domain specific partial ordering information. In fact, opposite to the method in~\cite{jiang2015self} which 
needs a full-order curriculum, our approach integrates prior knowledge in a form of partial curriculum. Thus, it can be 
applied to large scale (SBIR) tasks.

\subsection{Cross-domain Dictionary Learning}
Dictionary learning~\cite{lee2006efficient} is a popular method for finding effective sparse representations of input 
data. DL has been successfully applied in various image processing and computer vision tasks, such as image 
denoising~\cite{mairal2009non} and video event detection~\cite{yan2015complex}. With the fast emergence of large scale 
cross-domain datasets, traditional DL approaches have been extended to cross-modal tasks. For instance, Yang 
\etal\cite{yang2010image} proposed to learn a set of source-specific dictionaries from samples corresponding to 
different domains in a coupled fashion in the context of image super-resolution. In~\cite{wang2012semi} Wang \etal
introduced semi-coupled DL for photo-sketch synthesis, where source-specific dictionaries are learned together with a 
mapping function which describes the intrinsic relationship between domains. Similarly, Huang and Wang 
\cite{huang2013coupled} proposed a framework to simultaneously learn a pair of domain-specific dictionaries and the 
associated representations. Coupled DL approaches have also been applied to SBIR both in~\cite{wang2012semi} 
and~\cite{huang2013coupled} and to other related tasks, such as sketch-based 3D object retrieval~\cite{wang2015sketch} and sketch recognition~\cite{guo2015building}. However, none of these cross-domain DL methods explore self-paced learning or 
curriculum learning to construct more robust features.

\section{The Proposed Approach}\label{sec:method}
As discussed in Section \ref{intro}, in this paper we introduce a novel cross-domain representation learning framework 
for sketch-based image retrieval. Figure~\ref{framework} shows an overview of our approach. The overall objective of the 
proposed model is to learn robust cross-modal feature representations. As previously mentioned, commonly used 
cross-modal representation leaning methods, such as coupled dictionary learning~\cite{yang2010image} and multi-modal 
deep learning~\cite{ngiam2011multimodal}, usually rely on non-convex optimization problems and are likely to get 
stuck at a bad local optimal. We investigate how to incorporate the ideas of SPL and partial curriculum learning within a 
principled unified dictionary-based learning framework. 
%

In the following, we describe the proposed approach in details, presenting the general formulation of the overall learning 
problem (Section~\ref{subsec:ps}), the details of CPPCL (Section~\ref{subsec:CPRL}), the instantiation of CPPCL into CDL
(Section~\ref{CPPCLCDL}) and the construction of modality-specific curricula (Section~\ref{subsec:curriculum}). 

\subsection{Problem Formulation}
\label{subsec:ps}
Let us assume the existence of $K$ sketches and denote the features extracted from the $k$-th sketch as $\vect{f}_k^\sketch\in\mathbb{R}^{m_\sketch}$. Similarly, we assume the existence of $L$ images and denote the features extracted from the 
$l$-th image as $\vect{f}_l^\image\in\mathbb{R}^{m_\image}$. 
Each sketch (resp. image) corresponds to a new cross-modal representation to be learned, denoted as $\vect{c}^\sketch_k\in\mathbb{R}^N$ (resp. $\vect{c}^\image_l\in\mathbb{R}^N$) with $N$ being the dimension of the new representation. We also define $\vect{F}^\sketch = 
[\vect{f}_1^\sketch,\ldots,\vect{f}_K^\sketch]\in\mathbb{R}^{m_\sketch \times K}$ as the matrix of all sketch features, and $\vect{F}^\image$, 
$\vect{C}^\sketch$ and $\vect{C}^\image$ 
analogously. We denote $\mathcal{U}^\sketch$ and $\mathcal{U}^\image$ as the modality-specific partial curricula constructed from the sketch and the image domains respectively. The overall learning objective of the proposed cross-paced representation learning with partial curricula model can be written as:
\begin{equation}
\begin{aligned}
\min_{\vect{C}^\sketch, \vect{C}^\image, \vect{V}^\joint, \gvect{\xi}^\joint} &\,\,
\mathcal{L}_{\textrm{RL}}(\vect{C}^\sketch, \vect{C}^\image,\vect{V}^\joint; \vect{F}^\sketch, \vect{F}^\image) \\ &\!+\!  
f_{\textrm{PC}}(\gvect{\xi}^\joint; \mathcal{U}^{\sketch}, \mathcal{U}^{\image}) \!+\! f_{\textrm{SP}}(\vect{V}^\joint; \gamma) \\
\textrm{s.t.} \quad \quad &\,\,  v_k^\sketch,v_l^\image \in \{0, 1\} \quad \forall k,l
\end{aligned}
 \label{eq.overallobj}
\end{equation}
where $\vect{V}^\joint=\textrm{diag}(\vect{V}^\sketch, \vect{V}^\image)$ with $\vect{V}^\sketch = \textrm{diag}(v_1^\sketch, ..., v_K^\sketch)$ and $\vect{V}^\image = \textrm{diag}(v_1^\image, ..., v_L^\image)$, are binary pacing variables which indicate whether a training instance (sketch or image) has to be used for learning or not. $\mathcal{L}_{\textrm{RL}}(\vect{C}^\sketch, \vect{C}^\image,\vect{V}^\joint; \vect{F}^\sketch, \vect{F}^\image)$ is a cross-modal representation learning term given $\vect{F}^\sketch$ and $\vect{F}^\image$. For the proposed learning framework, this term is flexible to employ various representation learning methods such as coupled dictionary learning~\cite{huang2013coupled}, cross-domain subspace learning~\cite{wang2013learning} and deep learning~\cite{feng2014cross}.  $f_{\textrm{SP}}(\vect{V}^\joint; \gamma)$ is a cross-modal self-paced regularizer determining the learning order of samples in two modalities, and $\gamma \geq 0$ is a self-paced parameter which controls the 
learning pace. $f_{\textrm{PC}}(\gvect{\xi}^\joint; \mathcal{U}^{\sketch}, \mathcal{U}^{\image})$ is a partial curriculum (PC) regularizer which makes the learning order match with the pre-determined modality-specific curricula $ \mathcal{U}^{\sketch}$ and $\mathcal{U}^{\image}$ as much as possible, and $\gvect{\xi}^\joint$ represent partial curriculum learning variables. In the following, we present the details of the proposed learning framework. 

\subsection{Cross-paced Partial Curriculum Learning}
\label{subsec:CPRL}

CPPCL is a joint learning paradigm which combines a self-paced and a partial curriculum learning scheme, 
corresponding to the two components $f_{\textrm{PC}}(\gvect{\xi}^\joint; \mathcal{U}^{\sketch}, \mathcal{U}^{\image})$ and 
$f_{\textrm{SP}}(\vect{V}^\joint; \gamma)$ as described in Eqn.~\ref{eq.overallobj}. By doing so, the learning order is 
simultaneously determined by the pre-defined prior knowledge (\ie~partial-order modality-specific curriculum) and the
feedback from the learner during training. 

As mentioned in Section~\ref{subsec:ps}, in the self-paced learning philosophy, there is a pacing binary 
variable $v^\sketch_k\in\{0,1\}$ (respectively $v^\image_l\in\{0,1\}$) associated to sketch $k$ (respectively to image $l$), 
determining the learning order of the training samples. Importantly, $v^\sketch_k$ and $v^\image_l$ are not fixed and evolve 
during the training phase. To analyze the influence of the self-paced learning scheme, we investigate two different self-paced regularizers 
in our learning framework. 

\subsubsection{Self-paced regularizer A} is proposed to take into account the diversity of training data. We assume 
that the training data of the sketch modality are split into $G^\sketch$ groups or classes (either learned from the data 
or provided in advance). We define a group-specific indicator vector $\vect{p}_i^\sketch\in\mathbb{R}^{K}$, 
where $p_{i,k}^\sketch=1$ if and only if sample $k$ belongs to group $i$ ($i\in\{1,...,G^\sketch\}$), and $p_{i,k}=0$ 
otherwise. We devise a penalty over $\vect{V}^\sketch$ that is normalized over the groups' size, denoted by 
$E_i^\sketch$. The definitions in the image domain, \ie for $G^\image$, $\vect{p}_j^\image$ and $E_j^\image$ are 
analogous. The regularizer writes:
\begin{equation}
 f_{\textrm{SP}_A}(\vect{V}^\joint; \gamma) = -\gamma \left(\sum_{i=1}^{G^\sketch} \frac{1}{E_i^\sketch}\|\vect{V}^\sketch\vect{p}_i^\sketch\|_1 + \sum_{j=1}^{G^\image} \frac{1}{E_j^\image}\|\vect{V}^\image\vect{p}_j^\image\|_1 \right).
\end{equation}

This term enforces learning from different groups/classes and therefore it is closely related to SPL with 
diversity~\cite{jiang2014self}. Similarly to \cite{jiang2014self}, the idea is to learn not only from easy samples as in the standard SPL~\cite{kumar2010self} but also from samples that are dissimilar from what has already been learned. However, 
with respect to~\cite{jiang2014self}, the proposed regularizer has two prominent advantages: (i) we avoid using group 
norms that significantly increase the complexity of the optimization solvers and (ii) we introduce the normalization 
factors $E_i^\sketch$ and $E_j^\image$ that soften the bias induced by dissimilar group cardinalities.

\subsubsection{Self-paced regularizer B} introduces a slight modeling change. Indeed, following 
Zhao~\etal~\cite{zhao2015self} we consider the self-pacing variables $v_k^\sketch$ and $v_l^\image$ to be continuous in 
the range $[0,1]$. With this choice, we allow the model to take a soft decision and assess the importance of the 
training sample, rather than force the method to choose between using/ignoring the sample at the current iteration. 
Notice that the previous self-paced regularizer ($f_{\textrm{SP}_A}$) can also be used with continuous self-pacing 
variables. In addition, considering $v_k^\sketch$ and $v_l^\image$ to be continuous opens the door to the definition of 
more sophisticated self-pacing regularizers such as:
\begin{equation}
 f_{\textrm{SP}_B}(\vect{V}^\joint; \gamma) = - \frac{\gamma}{2} \left(\sum_{k=1}^{K} Q(v_k^\sketch) +  \sum_{l=1}^{L} Q(v_l^\image) \right),
 \label{eq.r}
\end{equation}
where $Q(v)= v^2-2v$ as in~\cite{zhao2015self}.

Importantly, the penalty induced by the regularizer evolves over time so as to incorporate more and more samples to be 
part of the training set. Specifically, the self-paced parameter $\gamma$ is multiplied by a step size $\eta$ ($\eta > 
1$) in order to increase $\gamma$ at each iteration, as in traditional SPL methods~\cite{kumar2010self}. This is done 
for both $f_{\textrm{SP}_A}$ and $f_{\textrm{SP}_B}$ regularizers.

An important methodological contribution of our work is to include \textbf{modality-specific partial curricula} into a representation learning framework and to study its behavior within the SPL strategy already discussed. Subsequently, we assume the existence 
of two modality-specific sets of constraints ${\cal U}^\sketch$ and ${\cal U}^\image$. Each element of the sets consists of an index pair representing that if 
$(k,k')\in{\cal U}^\sketch$, then $v^\sketch_k<v^\sketch_{k'}$ and learning should be performed considering a priori $\vect{f}^\sketch_{k'}$ before $\vect{f}^\sketch_k$, as it corresponds to an easier sample. 
Depending on the way the curricula are constructed ${\cal U}^\sketch$ could contain incompatibilities, for instance, 
$\{(k,k'),(k',k''),(k'',k)\}\subset{\cal U}^\sketch$. In addition, the cross-modal terms could also induce incompatibilities between the two 
modalities. Therefore, it is desirable to relax the constraints using a set of slack variables $\xi_{kk'}^\sketch$, $\xi_{ll'}^\image$, and the partial curriculum regularizer is written as:
\begin{equation}
f_{\textrm{PC}}(\gvect{\xi}^\joint; \mathcal{U}^{\sketch}, \mathcal{U}^{\image}) = \mu \left(\sum_{(k,k')\in{\cal U}^\sketch} \xi_{kk'}^\sketch + \sum_{(l,l')\in{\cal U}^\image} 
\xi_{ll'}^\image\right),
\end{equation}
where $\gvect{\xi}^\joint=[[\xi_{kk'}^\sketch]_{(k,k')\in{\cal U}^\sketch}[\xi_{ll'}^\image]_{(l,l')\in{\cal U}^\image}]$ is the vector of all slack 
variables and $f_{\textrm{PC}}$ is the partial curricula regularizer regulated by the parameter $\mu\geq 0$.
In all, the optimization problem of CPPCL writes:
\begin{align}
\nonumber\min_{\vect{V}^\joint, \gvect{\xi}^\joint} &\,\,
f_{\textrm{PC}}(\gvect{\xi}^\joint; \mathcal{U}^{\sketch}, \mathcal{U}^{\image}) + f_{\textrm{SP}}(\vect{V}^\joint; \gamma) \\
\nonumber&\,\, v_k^\sketch,v_l^\image \in \{0, 1\} \quad \forall k,l,\\
\nonumber& \,\,v_k^\sketch - v_{k'}^\sketch < \xi_{kk'}^\sketch,\, \xi_{kk'}^\sketch\geq 0,\, \forall (k,k')\in{\cal U}^\sketch \\
\nonumber& \,\,v_l^\image - v_{l'}^\image < \xi_{ll'}^\image,\, \xi_{ll'}^\image\geq 0,\, \forall (l,l')\in{\cal U}^\image.
\end{align}

\subsection{Instantiation of CPPCL into CDL}
\label{CPPCLCDL} 
To learn cross-modal representations for SBIR, we embed the CPPCL into a coupled dictionary learning framework. Given the feature matrices 
$\vect{F}^\sketch$ and $\vect{F}^\image$ of the sketch and image domain, and two $N$-word dictionaries, one per modality: 
$\vect{D}^\sketch=[\vect{d}_n^\sketch]_{n=1}^N\in\mathbb{R}^{m_\sketch\times N}$ and 
$\vect{D}^\image=[\vect{d}_n^\image]_{n=1}^N\in\mathbb{R}^{m_\image\times N}$, we learn the 
associated dictionaries $\vect{D}^\sketch$, $\vect{D}^\image$ and sparse representations 
$\vect{C}^\sketch,\vect{C}^\image$ by minimizing the following objective function:
\begin{equation}
\begin{aligned}
&\mathcal{L}_{\textrm{RL}} = 
\|(\vect{F}^\sketch-\vect{D}^\sketch \vect{C}^\sketch)\vect{V}^\sketch\|^2_\mathcal{F} + 
\|(\vect{F}^\image-\vect{D}^\image \vect{C}^\image)\vect{V}^\image\|^2_\mathcal{F}  \\ & 
+ \alpha\left( \|\vect{C}^\sketch\|_1 + \|\vect{C}^\image\|_1 
\right) + 
\beta\,\textrm{Tr}\left(\vect{C}^\joint\vect{V}^\joint\vect{L}\vect{V}^{\joint\top}\vect{C}^{\joint\top}\right),
\nonumber
\label{eq.dr.no-sp}
\end{aligned}
\end{equation}
subject to:
\begin{equation*}
 \|\vect{d}_n^\sketch\|,\|\vect{d}_n^\image\|\leq 1 \quad \forall n, \qquad v_k^\sketch,v_l^\image \in \{0, 1\} 
\quad \forall k,l,
\end{equation*}
where $\alpha\geq0$ is a regularization parameter and $\|\cdot\|_\mathcal{F}$ denotes the Frobenius norm. 
The constraints remove any scale ambiguities due to the matrix products $\vect{D}^\sketch \vect{C}^\sketch$ and $\vect{D}^\image \vect{C}^\image$,
while the regularization terms induce sparsity in the learned codes.

We also introduce a graph Laplacian regularizer to maintain the relational link between the learned representations of 
sketches and images in the training set. Ideally, each sketch corresponds to at least an image (\eg for sketch to photo 
face recognition~\cite{WangT09} in the context of security and biometrics applications). Alternatively, the association 
among sketches and images is derived from image class information~\cite{hu2013performance}.  Generally speaking, in this 
paper we consider both intra-modality and cross-modality relationships, modeled by a non-negative weight matrix 
$\vect{W}\in\mathbb{R}^{+\,(K+L)\times(K+L)}$. Intuitively, the larger $w_{pq}$ is, the stronger the relationship 
between the $p$-th and $q$-th codes is. Importantly, when $1\leq p,q\leq K$ (respectively, $K<p,q\leq K+L$), $w_{pq}$ 
relates two sketches (respectively, two images) creating an intra-modality link, otherwise $w_{pq}$ relates a sketch 
and an image (cross-modality link). Interpreting $\vect{W}$ as the weight matrix of a graph and denoting the associated 
Laplacian matrix\footnote{The Laplacian matrix of a graph with weight matrix $\vect{W}$ is defined as 
$\vect{L}=\vect{D}-\vect{W}$, where $\vect{D}$ is a diagonal matrix with $d_{pp} = \sum_{q}w_{pq}$.} by $\vect{L}$, a 
graph laplacian regularizer for the codes is defined as 
$\textrm{Tr}\left(\vect{C}^\joint\vect{L}\vect{C}^{\joint\top}\right) = \frac{1}{2}
\sum_{p,q=1}^{K+L}w_{pq}\|\vect{c}^\joint_p-\vect{c}^\joint_q\|^2$,
where $\vect{C}^\joint=[\vect{c}^\joint_p]_{p=1}^{K+L}=[\vect{C}^\sketch\,\vect{C}^\image]\in\mathbb{R}^{N\times(K+L)}$ 
is a joint code matrix, and 
$\beta\geq0$ is a regularization parameter controlling the importance of the relational knowledge. By embedding pacing variables $\vect{V}^\joint$ into $\textrm{Tr}\left(\vect{C}^\joint\vect{L}\vect{C}^{\joint\top}\right)$, we obtain the self-paced graph laplacian regularizer $\textrm{Tr}\left(\vect{C}^\joint\vect{V}^\joint\vect{L}\vect{V}^{\joint\top}\vect{C}^{\joint\top}\right)$.
Finally, the optimization problem to solve for writes:
\begin{equation}
\begin{aligned}
\min_{\vect{D}^\sketch,\vect{D}^\image,\vect{C}^\joint, \vect{V}^\joint, \vect{\xi}^\joint} &\,\,
\mathcal{L}_{\textrm{RL}} \!+\!  
f_{\textrm{PC}}(\gvect{\xi}^\joint; \mathcal{U}^{\sketch}, \mathcal{U}^{\image}) \!+\! f_{\textrm{SP}}(\vect{V}^\joint, \gamma) \\
\textrm{s.t.} & \,\,\|\vect{d}_n^\sketch\|,\|\vect{d}_n^\image\|\leq 1 \quad \forall n, \\
&\,\, v_k^\sketch,v_l^\image \in \{0, 1\} \quad \forall k,l, \\
& \,\,v_k^\sketch - v_{k'}^\sketch < \xi_{kk'}^\sketch,\, \xi_{kk'}^\sketch\geq 0,\, \forall (k,k')\in{\cal U}^\sketch \\
& \,\,v_l^\image - v_{l'}^\image < \xi_{ll'}^\image,\, \xi_{ll'}^\image\geq 0,\, \forall (l,l')\in{\cal U}^\image.
\end{aligned}
 \label{eq.overallobj1}
\end{equation}

\subsection{Laplacian and Curricula Construction}
\label{subsec:curriculum}
In this section we describe how we construct the modality-specific partial curricula and the Laplacian matrix 
representing the relational knowledge. However, it is worth noting that our approach is general and other design choices are possible. 
We build both the curricula and the Laplacian in the training set from the sketch and image features
and a group association, that could arise from the class membership or from unsupervised clustering. In our experiments, we also devised 
a protocol to construct a curriculum for sketches from human manual annotations.

\subsubsection{Construction of graph Laplacian matrix} To build the Laplacian matrix (computed from the weights 
$w_{pq}$), the intra-modality 
relationships are 
defined using the Gaussian kernel and the inter-modality with group association, as in~\cite{zhai2013heterogeneous}:
\begin{equation}
{w_{pq}} = \left\{ \begin{array}{l}
\textrm{e}^{-\left\| \vect{f}_p^\sketch - \vect{f}_q^\sketch \right\|_2^2/2{\sigma ^2}}, \quad  p,q\leq K\\
\textrm{e}^{-\left\| \vect{f}_{p-K}^\image - \vect{f}_{q-K}^\image \right\|_2^2/2{\sigma ^2}}, \quad  K < p,q\\
1, \quad \begin{array}{l} p\leq K < q \text{ and $p\sim q$}\\
	 q\leq K < p \text{ and $q\sim p$} \end{array} \\
0,  \quad \text{otherwise,}
\end{array} \right.
\end{equation}
where $\sigma$ is the Gaussian kernel parameter fixed to $1$ with no significant performance variation around this value. The symbol $\sim$ 
indicates samples belonging to the same cluster/class.

\subsubsection{Construction of modality-specific curricula}
Regarding the curricula construction, as stated above, a fundamental aspect of the the proposed framework is the possibility to 
handle partial curricula. Previous CL or hybrid CL-SPL methods~\cite{bengio2009curriculum,jiang2015self} instead assume that a 
full curriculum, \ie a complete order of samples, is provided. This is a strong assumption that may be unrealistic in 
real-world large-scale tasks. On the one hand, even if automatic measures of the easiness of an 
image~\cite{lee2011learning} have been developed, these metrics are accurate up to some extent and therefore deriving 
a full ranking from these measures may be inappropriate. On the other hand, manually annotating the entire set of 
images represents a huge human workload, highly demanding for medium and large-scale datasets. In addition, if the 
multi-modal dataset is gathered incrementally, the cost of updating the curriculum grows with the size of the dataset.
%

\begin{figure}[!t]
  \centering
  \includegraphics[width=0.478\textwidth]{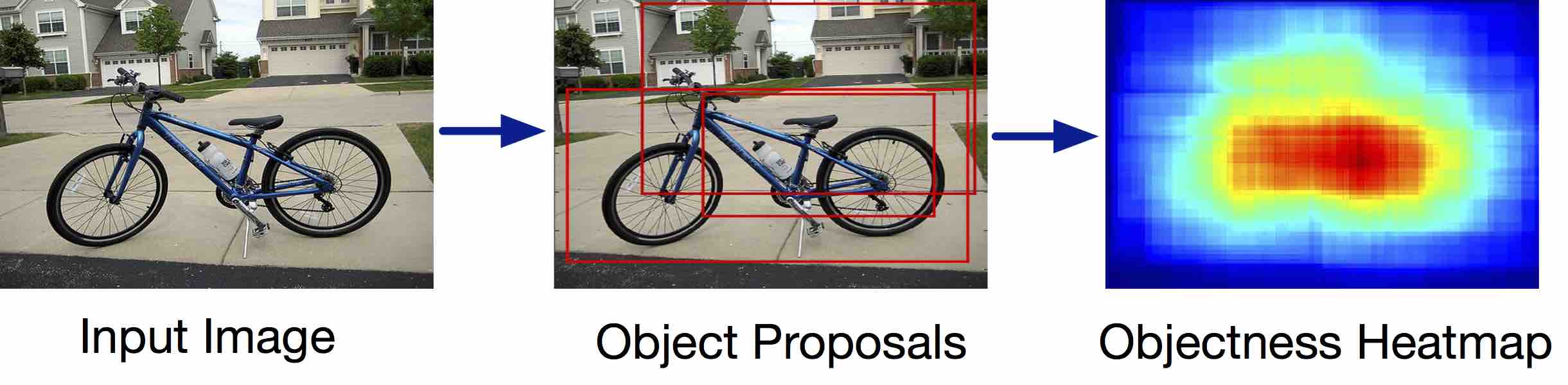} 
  \caption{An illustration of objectness generation process for assessing the easiness of an image sample.}
  \label{objectness}
  \vspace{-13pt}
\end{figure}

\paragraph{Image partial curricula} The partial curricula for the image domain is obtained by means of an automated 
procedure based on previous studies~\cite{lee2011learning,alexe2012measuring}. Intuitively, easy images are those 
containing non-occluded high-resolution objects in low-cluttered background. Previous works~\cite{lee2011learning} 
proposed to define the easiness of an image from the ``objectness'' measures~\cite{alexe2012measuring}. In the same line 
of though, we compute the easiness measure as the median of the 30 highest ``objectness'' scores among a set of 1,000 
window proposals. An example is shown in Figure~\ref{objectness}. This procedure approximates the easiness of 
a training image. Notice that, two images with largely different scores are likely to correspond to samples with 
different easiness. On the contrary, if the scores are similar, imposing that the image with the lowest score is the 
easiest in the pair may induce some errors. The constraint associated to an image pair is included in ${\cal U}^\image$ 
only if the difference of their associated scores exceeds a certain threshold $\delta^\image$ (\ie~if one of the images 
in the pair is significantly easier than the other).

\begin{figure}[!t]
  \centering
  \includegraphics[width=0.478\textwidth]{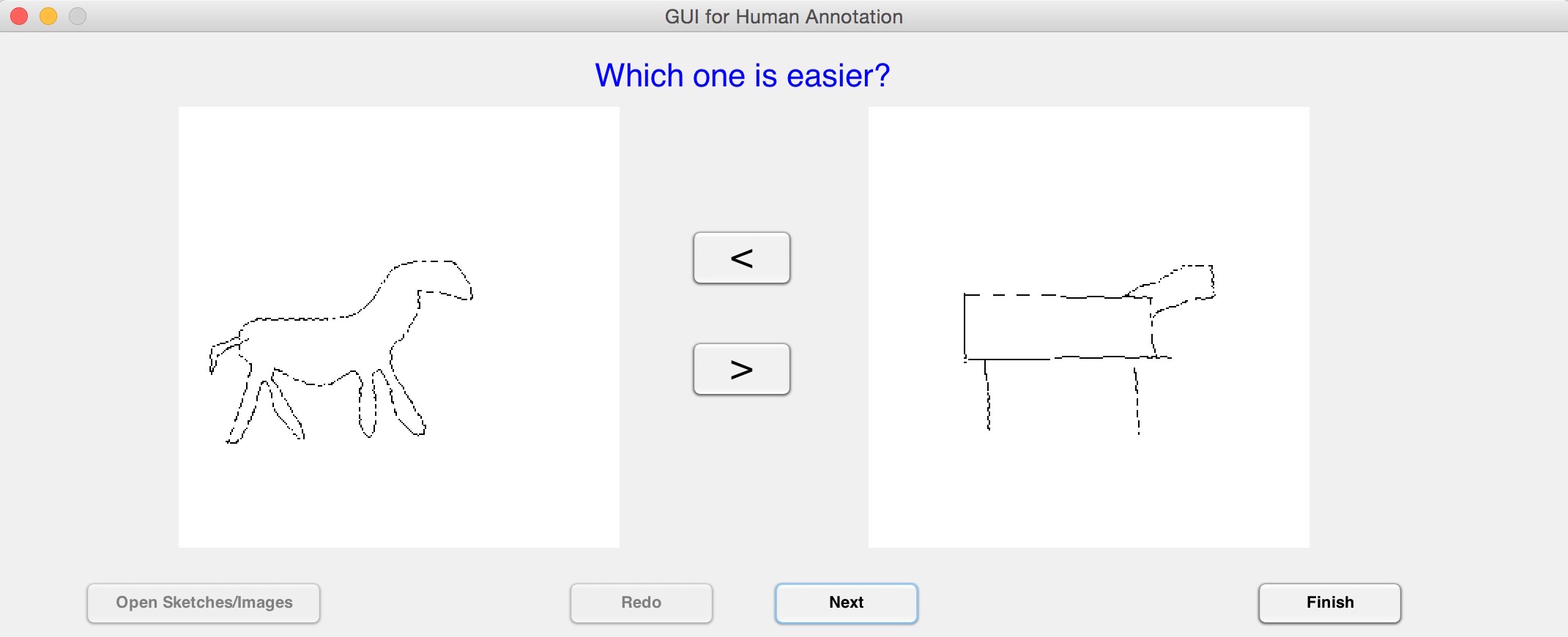} 
  \caption{The graphical interface used for annotation. Easy sketches are those with more details and easy images are 
  those with non-occluded high-resolution objects in low-cluttered background.}
  \label{GUI}
    \vspace{-13pt}
\end{figure}

\paragraph{Sketch partial curricula} Contrary to the image domain, there is no widely-accepted procedure to define 
the easiness of a sketch. Therefore, we consider two methods for constructing the partial curriculum in the sketch 
domain. The first one is an automatic method that follows again the philosophy of~\cite{alexe2012measuring}. Given a sketch, we 
randomly sample 100 windows at different scales and positions. For each window we compute the ``edgeness'' score, 
representing the edge density within the window as proposed in~\cite{alexe2012measuring}. Intuitively, the edgeness 
should follow the rationale that easy sketches are those with more details. As previously done for images, the 
constraint associated to a pair of sketches is included into ${\cal U}^\sketch$ only if their measure of edgeness 
differs by at least $\delta^\sketch$. The second one is a semi-automatic strategy for building the partial curricula of the 
sketches by including human annotators in the loop. A naive retrieval method based on SHOG 
features~\cite{eitz2011sketch} generates potential constraints (pairs of sketches). In details, we pair each sketch with 
the closest among the cluster/class. The human annotator is then queried which sketch is easier to learn from. Ten PhD 
students (6 male, 4 female) of age $24.3\pm1.4$ (mean, standard deviation) performed the annotation after being 
instructed that ``easy'' sketches meant sketches with more details. Importantly, since CPPCL is specifically designed to 
handle partial curricula, annotators had the possibility to ``skip'' sketch pairs if they were unable to decide. A 
simple GUI, shown in Figure~\ref{GUI}, was developed for annotation.

\section{Model Optimization}
\label{sec:optim}
\label{subsec:solver}
The optimization problem in Eqn.~\ref{eq.overallobj1} is not jointly convex in all variables. However, efficient 
alternate optimization techniques can solve it since it is convex on $\{\vect{D}^\sketch,\vect{D}^\image\}$, 
$\{\vect{C}^\joint\}$ and $\{\vect{V}^\joint,\gvect{\xi}^\joint\}$ when the other two sets of variables are fixed. We 
proposed two different self-paced regularizers $f_{\textrm{SP}_A}$ and $f_{\textrm{SP}_B}$ in our model. However, they 
have no impact when solving for $\vect{D}^\sketch$ and $\vect{D}^\image$, while we provide two different solutions for 
solving $\vect{V}^\joint$ and $\gvect{\xi}^\joint$.
\setcounter{subsubsection}{0}
\subsubsection{Solve for $\vect{D}^\sketch$ and $\vect{D}^\image$} Fixing $\vect{C}^\joint$, $\vect{V}^\joint$ and 
$\gvect{\xi}^\joint$,
the optimization problem for $\vect{D}^\sketch$ (analogously for $\vect{D}^\image$) writes:
\begin{equation}
\mathop {\min }\limits_{{\vect{D}^\sketch}} {\left\| {({\vect{F}^\sketch} - {\vect{D}^\sketch}{\vect{C}^\sketch})\vect{V}^\sketch} 
\right\|_2^2} \quad \textrm{s.t. } \left\| {{\vect{d}^\sketch_k}} \right\| \le 1.
\label{eq.sol.d}
\end{equation}
This problem is a Quadratically Constrained Quadratic Program (QCQP) that can be solved using gradient descent with e.g. Lagrangian 
duality~\cite{lee2006efficient}.

\subsubsection{Solve for $\vect{C}^\joint$} By fixing $\vect{D}^\sketch$, $\vect{D}^\image$, $\vect{V}^\joint$ and 
$\gvect{\xi}^\joint$ the 
optimization 
function for the codes can be rewritten as:
\begin{align}
\nonumber f(\vect{C}^\joint) &=  \left\| \left( \vect{F}^\sketch - \vect{D}^\sketch\vect{C}^\sketch \right)\vect{V}^\sketch \right\|_{\cal F}^2  + 
\left\| \left( \vect{F}^\image - \vect{D}^\image\vect{C}^\image \right)\vect{V}^\image \right\|_{\cal F}^2 \\
& + \alpha \left\| \vect{C}^\joint \right\|_1 + 
\beta\,\textrm{Tr}\left(\vect{C}^\joint\vect{V}^\joint\vect{L}\vect{V}^{\joint\top}\vect{C}^{\joint\top}\right).\label{eq.sol.c} 
\end{align}

According to FISTA~\cite{BeckT09}, $f$ can be viewed as a proximal regularization problem, solved using the following 
recursion (over $r$):
\begin{equation}
\vect{C}^\joint_r \!=\! \mathop {\arg\!\min }\limits_{{\vect{C}^\joint}} \!\left\{\frac{\left\| {{\vect{C}^\joint} \!-\! 
\vect{C}^\joint_{r 
- 1} \!+\! {t_r}\nabla\! f\!(\vect{C}^\joint_{r - 1})} \right\|_{\cal F}^2}{2 t_r}  + {\alpha}{\left\| {{\vect{C}^\joint}} 
\right\|_1}\right\},\label{eq.s3}
\end{equation}
where $t_r > 0$ is the step size and $\nabla\!f(\vect{C}^\joint) = [\nabla\!f(\vect{C}^\sketch) \; \nabla\!f(\vect{C}^\image)]$ is the 
concatenation of the two gradients defined as:
\begin{align}
 \nonumber \nabla\!f(\vect{C}^\sketch) &=  2\vect{D}^{\sketch\top}(\vect{D}^\sketch\vect{C}^\sketch - \vect{F}^\sketch)(\vect{V}^\sketch)^2 \\
 &+ 2\beta\left( \vect{C}^\sketch \vect{V}^\sketch \vect{L}^\sketch + \vect{C}^\image \vect{V}^\image \vect{L}^{\image\sketch} 
\right)\vect{V}^\sketch,
\label{eq.sol.s2}
\end{align}
where the sublaplacian matrices are taken from the Laplacian matrix as $\vect{L} = [\vect{L}^\sketch \, \vect{L}^{\sketch\image} ; 
\vect{L}^{\image\sketch} \, \vect{L}^\image]$. The second gradient, $ \nabla\!f(\vect{C}^\image)$ is defined analogously to $ 
\nabla\!f(\vect{C}^\sketch)$. Moreover, (\ref{eq.s3}) is a standard LASSO problem whose optimal solution can be found using the feature-sign search 
algorithm in~\cite{lee2006efficient}.
  
\subsubsection{Solve for $\vect{V}^\joint$ and $\gvect{\xi}^\joint$ with the regularizer $f_{\textrm{SP}_A}$} We fix 
$\vect{D}^\sketch$, $\vect{D}^\image$, $\vect{C}^\joint$ to solve for $\gvect{\xi}^\joint$ and $\vect{V}^\joint$, and 
the problem writes:
\begin{equation}
\begin{aligned}
\min_{\vect{V}^\joint} \,\,  &\mathcal{L}_{\textrm{RL}}
-\gamma \left(\sum_{k=1}^K \frac{1}{E_{g,k}^\sketch}v_k^\sketch + \sum_{l=1}^L \frac{1}{E_{g', l}^\image}v_l^\image\right), \\
& + \mu\left(\sum_{(k,k')\in{\cal U}^\sketch} \xi_{kk'}^\sketch + \sum_{(l,l')\in{\cal U}^\image} 
\xi_{ll'}^\image\right) \\
\textrm{s.t.} \,\,  &0 \leq v_k^\sketch,v_l^\image \leq 1 \quad \forall k,l, \\
&   v_k^\sketch - v_{k'}^\sketch < \xi_{kk'}^\sketch,\, \xi_{kk'}^\sketch\geq 0,\, \forall (k,k')\in{\cal U}^\sketch, \\
&  v_l^\image - v_{l'}^\image < \xi_{ll'}^\image,\, \xi_{ll'}^\image\geq 0,\, \forall (l,l')\in{\cal U}^\image.
\label{eq.spb_v}
\end{aligned}
\end{equation}
Here we replace the self-paced regularizer $\sum_{g=1}^{G^\sketch} 
\frac{1}{E_g^\sketch}\|\vect{V}^\sketch\vect{p}_g^\sketch\|_1$ with $\sum_{k=1}^K \sum_{g=1}^{G^\sketch} 
\frac{1}{E_g^\sketch}p_{g, k}^\sketch v_k^\sketch = \sum_{k=1}^K \frac{1}{E_{g,k}^\sketch}v_k^\sketch$ since $v_i\geq 
0$, $E_{g,k}^\sketch$ being the size of group/class $g$ of sample $k$. As discussed in Section~\ref{subsec:CPRL} and following~\cite{zhao2015self, xumulti}, the self-paced regularizer can be used with continuous self-pacing variables for facilitating the optimization. This property is particularly advantageous in our case, because the joint $(\vect{V}^\joint,\gvect{\xi}^\joint)$ optimization problem can now be treated as a quadratic programming (QP) problem with a set of linear inequality constraints. Recent studies have shown that this strategy, as opposed to solving the original mixed integer quadratic programming problem, is successful 
in several applications~\cite{zhao2015self,jiang2015self}.

Let $\vect{y} = [[v_k^\sketch]_k [v_l^\image]_l [\xi^\sketch_{kk'}]_{kk'} 
[\xi^\image_{ll'}]_{ll'}]\in\mathbb{R}^{K+L+C^\sketch+C^\image}$ denote the joint optimization variable for which the 
problem writes:
%
%
\begin{align}
 \min_{\vect{y}} &\,\, \vect{y}^\top \vect{R} \vect{y} + \vect{b}^\top\vect{y} \label{eq.sol.v} \\
 \textrm{s.t.} &\,\, \vect{G}\vect{y} \leq \vect{h}, \nonumber
\end{align}
where the values of $\vect{R}$, $\vect{b}$, $\vect{G}$ and $\vect{h}$ are defined in the following. $\vect{R}$ is a 
$(K+L+C^\sketch+C^\image)\times(K+L+C^\sketch+C^\image)$ matrix with all zeros except for the first $(K+L)\times(K+L)$ block, where $C^\sketch$ and $C^\image$ denote the number of constraints of the sketch and the image modality respectively. More precisely:
\begin{equation}
 R_{pq} = \left\{\begin{array}{ll}
 \|\vect{f}_p^\sketch - \vect{D}^\sketch\vect{c}_p^\sketch\|^2 & q=p\leq K \\
 \|\vect{f}_{p-K}^\image - \vect{D}^\image\vect{c}_{p-K}^\image\|^2 & K<q=p\leq L+K \\
 \beta w_{pq} \|\vect{c}_p^\joint-\vect{c}_q^\joint\|^2 & 1\leq p\neq q \leq K+L \\
 0 & \text{otherwise}
 \end{array}\right. \nonumber
\end{equation}

and $\vect{b} = [-\frac{\gamma}{E_{g,1}^\sketch},...,-\frac{\gamma}{E_{g,K}^\sketch}, -\frac{\gamma}{E_{g',1}^\image},...,-\frac{\gamma}{E_{g',L}^\image}, \,\, \mu\vect{1}_{C^\sketch+C^\image}^\top]^\top$, where $\vect{1}_{C}$ is a $C \times 1$ vector filled with ones. $\vect{G}$ and $\vect{h}$ represent the inequality and bound constraints in (\ref{eq.overallobj1}) and their derivation is straightforward. Since there are $2(K+L+C^\sketch+C^\image)$ constraints, $\vect{G}\in\mathbb{R}^{2(K+L+C^\sketch+C^\image)\times(K+L+C^\sketch+C^\image)}$ and 
$\vect{h}\in\mathbb{R}^{2(K+L)+C^\sketch+C^\image}$.

\subsubsection{Solve for $\vect{V}^\joint$ and $\gvect{\xi}^\joint$ with the regularizer $f_{\textrm{SP}_B}$} Similar 
to the previous case with the regularizer $f_{\textrm{SP}_A}$, by fixing the dictionaries $\vect{D}^\sketch$, 
$\vect{D}^\image$ and the codes $\vect{C}^\joint$, the optimization problem is also a QP problem, and the only 
difference is that $R_{pq}$ and $\vect{b}$ change. In this case, $\vect{b} = [\gamma\vect{1}_{K+L}^\top \,\, 
\mu\vect{1}_{C^\sketch+C^\image}^\top]^\top$ and $R_{pq}$ becomes:
\begin{equation}
 R_{pq} = \left\{\begin{array}{ll}
 \|\vect{f}_p^\sketch - \vect{D}^\sketch\vect{c}_p^\sketch\|^2 - \gamma/2 & q=p\leq K \\
 \|\vect{f}_{p-K}^\image - \vect{D}^\image\vect{c}_{p-K}^\image\|^2 - \gamma/2 & K<q=p\leq L+K \\
 \beta w_{pq} \|\vect{c}_p^\joint-\vect{c}_q^\joint\|^2 & 1\leq p\neq q \leq K+L \\
 0 & \text{otherwise}
 \end{array}\right. \nonumber
\end{equation}
Then the QP problem can be effectively solved with the interior-point algorithms~\cite{vanderbei1999loqo}. The full 
optimization procedure is shown in Algorithm~\ref{algo:max}.

\begin{figure}[t]
 \centering
 \includegraphics[width=0.498\textwidth]{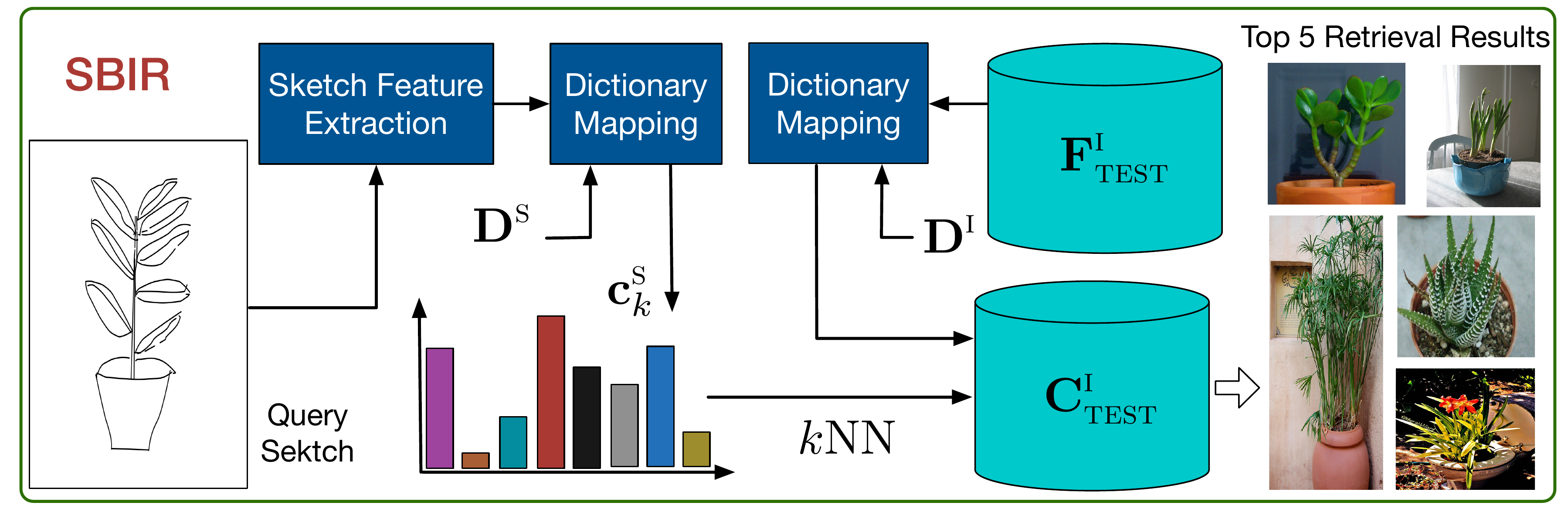} 
 \caption{An illustration of the test phase of the proposed model for SBIR.}
 \label{sbir_test_phase}
  \vspace{-5pt}
 \end{figure}
\begin{algorithm}[!t]
\DontPrintSemicolon 
\KwIn{the features $\vect{F}^\sketch$, $\vect{F}^\image$ and the parameters $\alpha$, $\beta$, $\gamma$, $\mu$}
\KwOut{$\vect{D}^\sketch$, $\vect{C}^\sketch$, $\vect{D}^\image$, $\vect{C}^\image$}
Initialize $\vect{D}^\sketch$, $\vect{C}^\sketch$, $\vect{D}^\image$, $\vect{C}^\image$ as described in 
Section~\ref{experiments} and initialize a step size $\eta$ ($\eta>1$);\\
\While{not converged} {
Update $\vect{V}^\joint$ and $\gvect{\xi}^\joint$ following~(\ref{eq.sol.v});\\
Update $\vect{C}^\sketch$, $\vect{C}^\image$ with~(\ref{eq.s3});\\
Update $\vect{D}^\sketch$, $\vect{D}^\image$ by solving~(\ref{eq.sol.d});\\
$\gamma \leftarrow \eta\gamma$; 
}
\Return{$\vect{D}^{\sketch^\star}$, $\vect{C}^{\sketch^\star}$, $\vect{D}^{\image^\star}$, $\vect{C}^{\image^\star}$}
\caption{Optimization Procedure }
\label{algo:max}
\end{algorithm} 

\subsubsection{Test Phase for Sketch-to-Image Retrieval}\label{testphase}
Fig.~\ref{sbir_test_phase} depicts the test phase of the proposed approach. 
Given the learned dictionaries $\vect{D}^\image$ and features of the retrieval images $\vect{F}^\image_{\textsc{TEST}}$, 
we perform a dictionary mapping to calculate all the sparse representations $\vect{C}^\image_{\textsc{TEST}}$ of the 
retrieval sketches via solving:
\begin{equation}
\mathop {\min }\limits_{{\vect{C}^\image_{\textsc{TEST}}}} {\left\| {({\vect{F}^\image_{\textsc{TEST}}} - {\vect{D}^\image}{\vect{C}^\image_{\textsc{TEST}}})} 
\right\|_{\cal{F}}^2} + \alpha \left\| {{\vect{C}^\image_{\textsc{TEST}}}} \right\|_1.
\label{eq.sol.test}
\end{equation}
For a query sketch $k$, a corresponding sparse representation $\vect{c}_k^\sketch$ can be calculated by a 
similar dictionary mapping with $\vect{D}^\sketch$ as in Eqn.~(\ref{eq.sol.test}). Then we retrieve top $K$ 
results from $\vect{C}^\image_{\textsc{TEST}}$ using $K$ Nearest Neighbor ($K$-NN), 
while for tests on sketch-to-face recognition, a Nearest Neighbor classifier is used.

\section{Experiments}\label{sec:exps}
\label{experiments}
To evaluate the effectiveness of our approach for Cross-Paced Representation Learning (CPRL), we conduct extensive experiments on four publicly available datasets: the CUHK Face Sketch 
(CUFS)~\cite{WangT09}, the Flickr15k~\cite{hu2013performance}, the Queen Mary SBIR~\cite{li2014fine} and the TU-Berlin Extension~\cite{eitz2012humans} datasets.
\subsection{Implementation Details}
The experiments were run on a PC with a quad core (2.1 GHz) CPU, 64GB RAM and an Nvidia Tesla K40 GPU. The proposed SBIR approach is implemented 
in Matlab and partially in C++ (the most computationally expensive components). For representing sketches, we adopted the mid-level 
representation method named Learned KeyShapes (LKS)~\cite{saavedrasketch}. We used a C++ implementation for efficient extraction of LKS features 
and wrap it in a Matlab interface. For representing images, CNN features were used. Specifically, the Caffe reference network `AlexNet' pre-trained 
on ImageNet was used to extract features from the sixth (the first fully connected) layer. In all our experiments and 
for all datasets, the value of the self-paced parameter was initialized to $\gamma = 1$ and increased by a factor $\eta = 1.3$ at each 
iteration (until all the training samples are selected). The dictionaries $\vect{D}^\sketch$ and $\vect{D}^\image$ 
were initialized with joint DL \cite{yang2010image} when both features have the same dimension and with modality-independent DL otherwise.

\subsection{Sketch-to-Face Recognition}

\begin{figure}[t]
\centering
\includegraphics[width=0.48\textwidth]{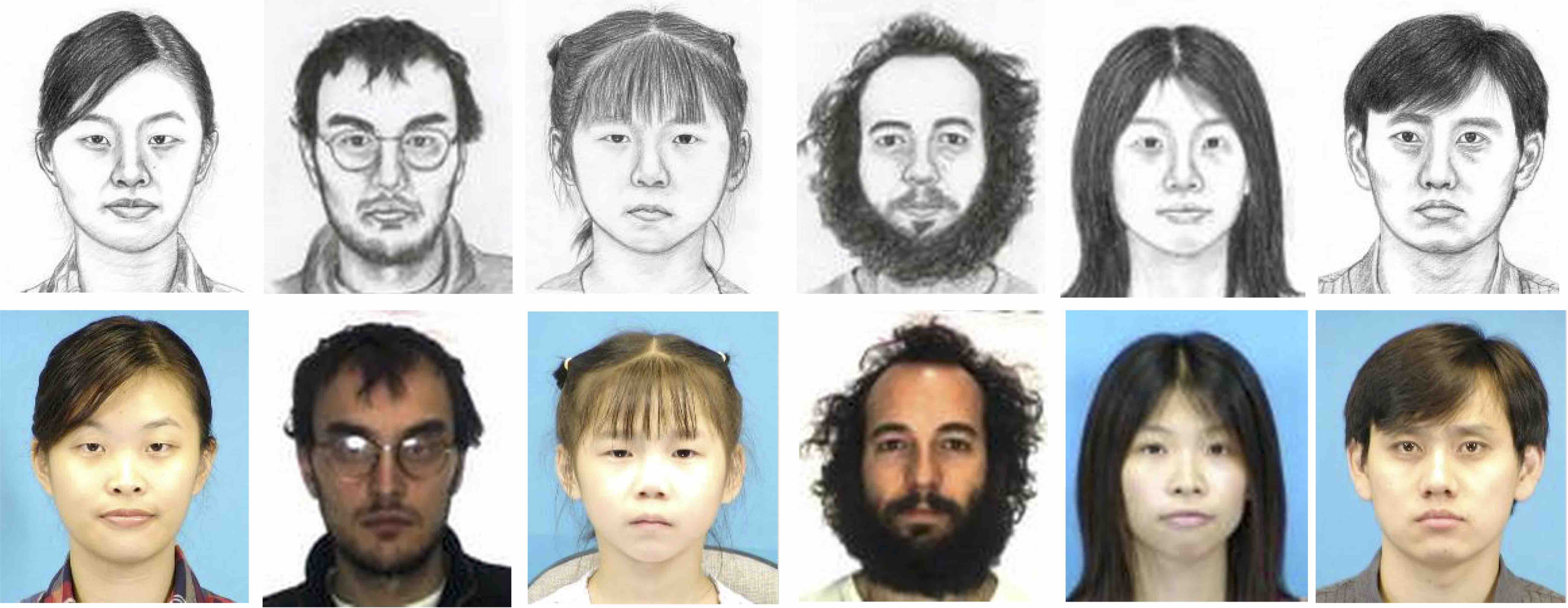}
\caption{Examples of sketch and face photo pairs of CUFS dataset.}
\label{cufsexamples}
\vspace{-13pt}
\end{figure}

\textbf{Dataset.} We first carried out experiments on sketch to face recognition using the \textbf{CUFS} dataset, a very popular benchmark 
which contains sketch-face photo pairs collected from 188 CUHK students. Figure~\ref{cufsexamples} shows some examples of sketch-photo pairs. 
The recognition task is to extract the 
face photo corresponding to a given sketch as described in Section~\ref{testphase}.  We evaluated the performance of our approach
on CUFS and compared it to other cross-domain retrieval methods and previous DL approaches.

\textbf{Settings.} Following~\cite{tang2004face}, in our experiments 88 sketch-photo image pairs were 
randomly selected for training the model, and the remaining 100 pairs were used for testing.
To fairly compare with previous works~\cite{yang2010image,huang2013coupled}, in this preliminary experiment 
we did not consider the powerful LKS sketch features and CNN image features, but we only used raw pixels as feature representations 
for the two modalities. 
We compared the proposed approach with several baseline methods including: canonical correlation 
analysis~(CCA), partial least squares (PLS)~\cite{sharma2011bypassing}, bilinear model~\cite{tenenbaum2000separating}, 
semi-coupled dictionary learning (SCDL)~\cite{wang2012semi}, joint dictionary learning (JDL)~\cite{yang2010image} and coupled dictionary learning 
(CDL)~\cite{huang2013coupled}. For the bilinear model, we used 70 PLS bases and 50 eigenvectors (see \cite{sharma2011bypassing}). For all 
DL-based approaches we set the dictionary size to 50.
In all cases, the recognition was performed using the nearest 
neighbor on the newly learned sparse representation as in~\cite{sharma2011bypassing,huang2013coupled}. 
We implemented and evaluated two variants of our method considering two different self-paced 
regularizers $f_{\textrm{SP}_A}$ and $f_{\textrm{SP}_B}$ as introduced in Section~\ref{subsec:CPRL}. Furthermore, we explicitly evaluated 
the importance of 
the relational knowledge ($\beta$) and of self-pacing ($\gamma$). Since for CUFS both sketches and face images are quite homogeneous (\ie, sketches 
were drawn by experts, faces in images are centered and equally illuminated), we did not use any curriculum by setting $\mu=0$. 
The parameters $\alpha$, $\beta$ were set by cross-validation to $1$ and $5$, respectively. 

\begin{table}[t]
\Large
\centering
\caption{Average recognition rate for all benchmarked methods on CUFS for sketch-to-photo face recognition.}
\resizebox{1.01\linewidth}{!} {
\begin{tabular}{lc}
\toprule
Method & \tabincell{c}{Recognition Rate} \\
\midrule
Tang \& Wang~\cite{tang2004face}  & 81.0\% \\
Partial Least Squares (PLS)~\cite{sharma2011bypassing}  &  93.6\% \\
Biliner model~\cite{tenenbaum2000separating} & 94.2\% \\
Canonical Correlation Analysis (CCA) & 94.6\% \\
Semi-coupled Dictionary Learning (SCDL)~\cite{wang2012semi}  & 95.2\% \\
Joint Dictionary Learning (JDL)~\cite{yang2010image} & 95.4\% \\
Coupled Dictionary Learning (CDL)~\cite{huang2013coupled}  & 97.4\% \\
\midrule
\midrule
CPRL with $f_{\textrm{SP}_B}$ ($\beta=\gamma = \mu = 0$) & 96.8\% \\ 
CPRL with $f_{\textrm{SP}_B}$ ($\gamma = \mu = 0$) & 97.2\% \\
CPRL with $f_{\textrm{SP}_A}$ ($\mu = 0$) & \textbf{98.2\%} \\
CPRL with $f_{\textrm{SP}_B}$ ($\mu = 0$) & \textbf{98.6}\% \\
\bottomrule
\end{tabular}
}
\label{sketch2photo}
  \vspace{-13pt}
\end{table}

\textbf{Results.} Table~\ref{sketch2photo} shows the results of average recognition rate over five trials. CPRL with self-paced 
regularizer $f_{\textrm{SP}_B}$ ($\mu=0$) achieves the best average recognition rate: $98.6\%$ (the influence of different self-paced 
regularizers is further analyzed in Section~\ref{subsec:deep-anal}). Remarkably, CPRL with $f_{\textrm{SP}_B}$($\mu=0$) outperforms CDL, 
which is the best of the DL-based approaches, showing the advantage of using our self-paced scheme for learning robust cross-domain representations. 
Importantly, by setting the parameter $\beta$ to 0, we notice that the 
effect of the relational knowledge is crucial in the performance of the overall method (CDL also uses relational knowledge). Among the compared 
methods, SCDL, JDL and CDL are the strongest competitors, achieving $95.2\%$, $95.4\%$ and $97.4\%$ recognition rate respectively. This means that 
DL is an effective strategy for learning cross-domain representations for the retrieval task. We also remark that CPRL 
with $f_{\textrm{SP}_B}$($\gamma=0$) outperforms the other two 
versions of CPRL, suggesting that the relational knowledge within the SP learning framework is beneficial for accurate retrieval.

\begin{figure*}[t]
\centering
\includegraphics[width=0.96\textwidth]{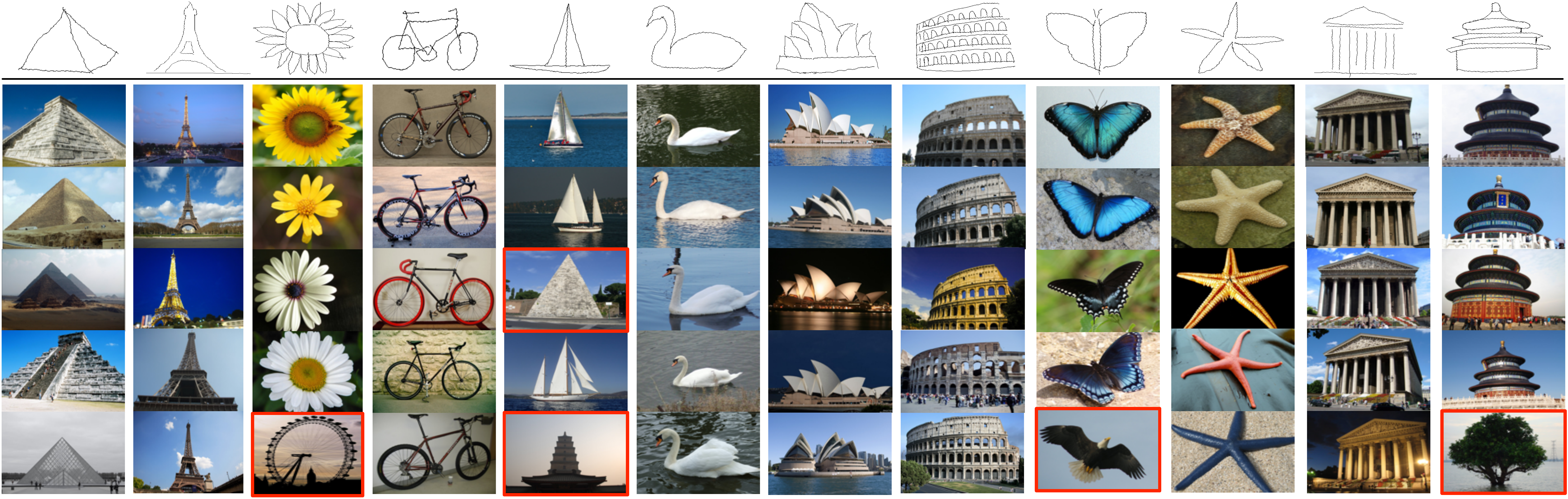} 
\caption{Top 5 retrieval results with sample query sketches in Flickr15K dataset. Red boxes show false positive retrievals.}
\label{flickqualitative}
  \vspace{-13pt}
\end{figure*} 
 
\begin{figure*}[t]
\centering
\subfigure[Flickr15K]{\includegraphics[width=0.445\textwidth, height=0.318\textwidth]{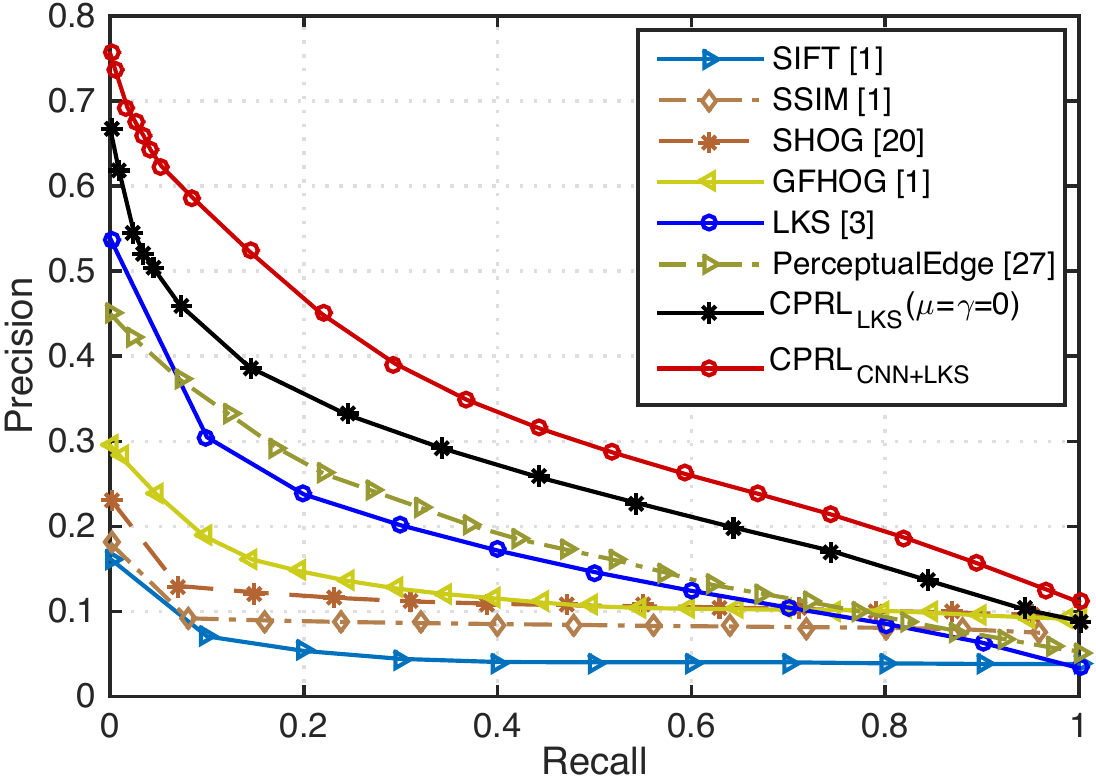}\label{flickrpr}}
\hspace{9pt}
\subfigure[QueenMary SBIR]{\includegraphics[width=0.445\textwidth, height=0.319\textwidth]{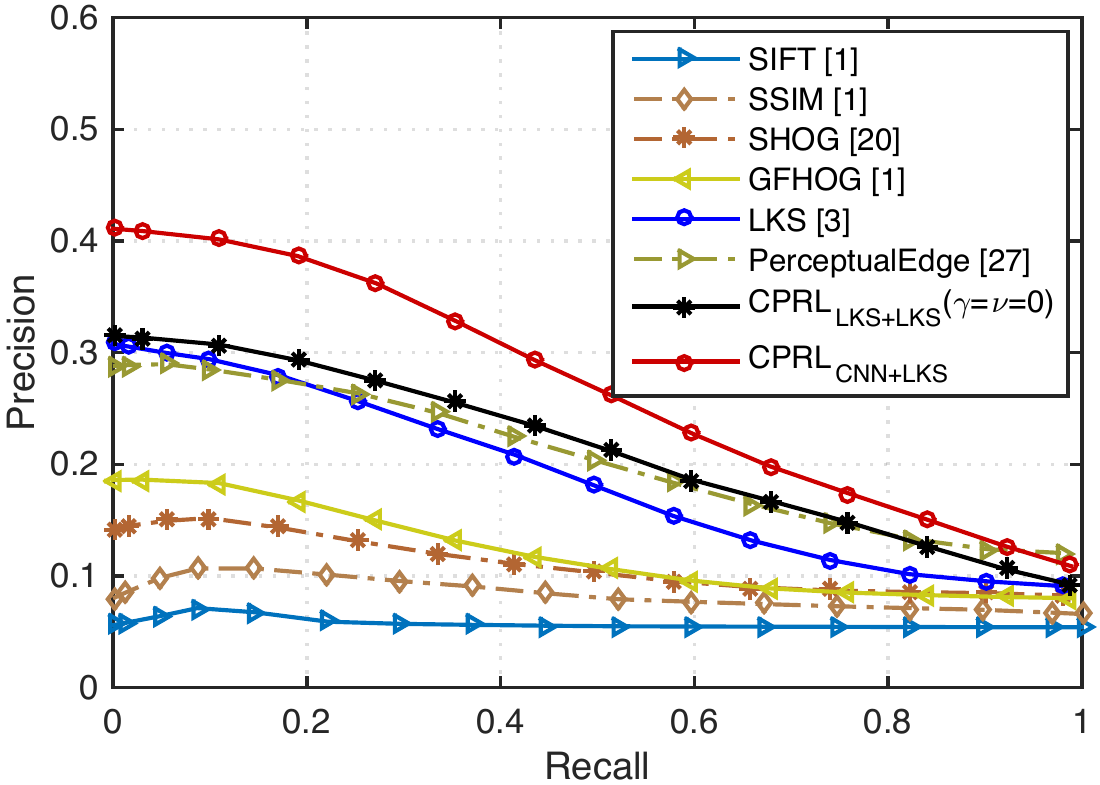}\label{queenpr}}
\caption{Precision-Recall (PR) curves for the retrieval performance comparison of the different methods on Flirckr15K and QueenMary SBIR datasets.}
\label{prcurve}
  \vspace{-13pt}
\end{figure*}

\subsection{Sketch-to-Image Retrieval}

\textbf{Datasets.} We further performed the evaluation of CPRL on the Flickr15k and QueenMary SBIR datasets. 
The \textbf{Flickr15k dataset} is a widely used dataset for SBIR, containing around 14,\,660 images collected from Flickr 
and 330 free-hand sketches drawn by 10 non-expert sketchers. The dataset consists of 33 object categories and each sample is labeled with an object-class annotation. Since this dataset does not provide a training set, to evaluate our approach, we partitioned the dataset into a training set with randomly chosen 40\% samples and a test set with the remaining samples. All the baseline methods were tested using the same setting for a fair comparison. 

The \textbf{Queen Mary SBIR dataset}~\cite{li2014fine} is constructed by intersecting 14 common categories from the Eitz 20,000 
sketch dataset~\cite{eitz2012hdhso} and the PASCAL VOC 2010 dataset~\cite{everingham2010pascal}, which consists of 1,120 sketches and 7,267 images. 
This dataset presents more complex conditions than the Flickr15k due to cluttered background and significant scale variations in the images. 
We use the official training and testing sets for evaluation. Since this dataset was originally used for fine-grained SBIR, while our 
task focuses on category-level SBIR, we only used image-level category annotations.

The \textbf{TU-Berlin Extension dataset}~\cite{eitz2012humans} consists of 250 object
categories and each category has 80 free-hand sketches. Similar to~\cite{liu2017deep}, 204,489 extended natural images provided by~\cite{zhang2016sketchnet} are added to TU-Berlin image gallery for the retrieval task.

\textbf{Settings.} To demonstrate the retrieval performance of CPRL, we compared with several state of the art SBIR methods, including 
SHOG~\cite{eitz2011sketch}, SIFT, SSIM, GFHOG evaluated in~\cite{hu2013performance}, Structure Tensor~\cite{eitz2010evaluation}, Learned Key 
Shapes as in~\cite{saavedrasketch}, PerceptualEdge~\cite{qi2015making}, Sketch-a-Net (SaN) \cite{yu2015sketch}, Siamese CNN~\cite{qi2016sketch}, GN Triplet~\cite{sangkloy2016sketchy}, 3D shape~\cite{wang2015sketch} and DSH~\cite{liu2017deep}. The first five methods first extract low-level feature representations (SHOG, SIFT, SSIM, GFHOG and 
StructureTensor) from the Canny edge maps of the images and the sketches respectively, and then generate the corresponding 
mid-level representations using a bag-of-words approach. Since the Queen Mary SBIR is a more difficult dataset than Flickr15, we 
considered SP-SHOG and SP-GFHOG instead of SHOG and GFHOG. Indeed, SP-HOG and SP-GFHOG employ a spatial pyramid model over SHOG and GFHOG features 
which has been demonstrated to provide more robust image 
representations than BoW~\cite{lazebnik2006beyond}. LKS~\cite{saavedrasketch} learns mid-level sketch patterns named keyshapes. The learned keyshapes are used to construct image and sketch descriptors. 
PerceptualEdge~\cite{qi2015making} uses 
an edge grouping framework to create synthesized sketches from images. The retrieval is performed by querying the 
synthesized sketches instead of the images directly. Sketch-a-Net (SaN)~\cite{yu2015sketch} is an approach based on recent CNN architectures. Siamese CNN~\cite{qi2016sketch} uses a Siamese-based network structure for learning the similarity between the image and the sketch samples. DSH~\cite{liu2017deep} jointly learns a hash function with the front-end CNN. 
For LKS and PerceptualEdge, we use the original codes provided by the authors 
with the same parameter setting described in the associated papers and we reimplemented other baselines whose codes are not publicly available. 
All the methods are evaluated on the same training/testing set for a fair comparison. If the original paper uses the same train/test split, the results are those reported in the paper.

To evaluate the proposed CPRL, we considered several settings using the self-paced regularizer $f_{\textrm{SP}_B}$: 
(i) CPRL$_\textsc{LKS}$ ($\gamma=\mu=0$): CPRL without the curricula and self-pacing, using LKS features for both image and sketch domains; 
(ii) CPRL$_\textsc{LKS}$: CPRL using LKS features for both the image and the sketch modalities; (iii) CPRL$_\textsc{CNN+LKS}$: CPRL using CNN features 
for the image domain (\ie features extracted from the sixth layer of the Caffe reference network trained on ImageNet) and LKS 
features for the sketch domain. We further considered the last baseline method with self-paced regularizer
$f_{\textrm{SP}_A}$. The sketch curriculum, when used, 
is constructed using $60\%$ of human annotations, since we did not observe any 
significant differences between the automatic and the manual procedures (see Section~\ref{subsec:deep-anal}). 
For all CPRL methods, we set $\alpha,\beta,\gamma$ and $N$ with cross-validation, and 
obtained $\alpha=2$, $\beta=25$, $\gamma=0.5$ and $N=1000$ for Flickr15k, and $\alpha=6$, $\beta=8$, $\gamma=1$ and $N=1500$ for Queen Mary SBIR.

\begin{figure*}[!t]
\centering
\includegraphics[width=0.999\textwidth]{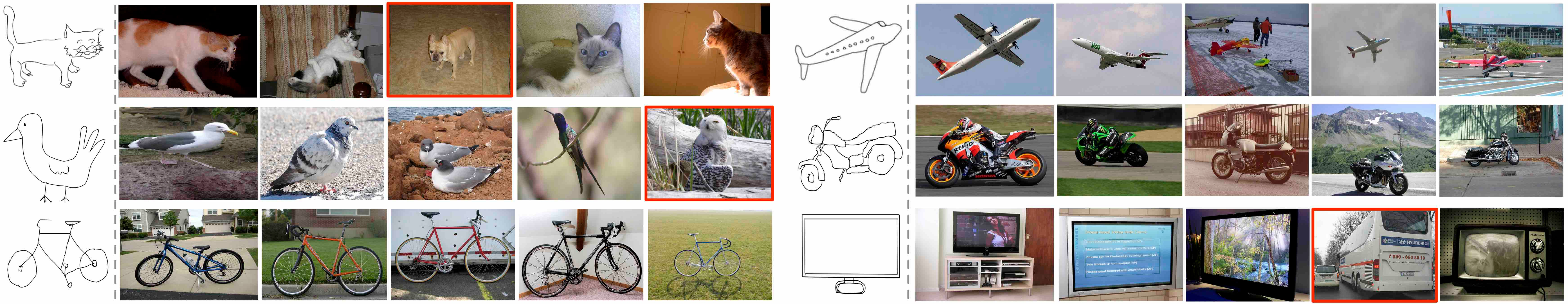}
\caption{Top 5 retrieved images (right) using the query sketch 
samples (left) in the QueenMary SBIR dataset. Red boxes show false positive retrievals.}
\label{queenmaryexample}
  \vspace{-13pt}
\end{figure*}

\begin{figure*}[!t]
\centering
\includegraphics[width=1.\textwidth]{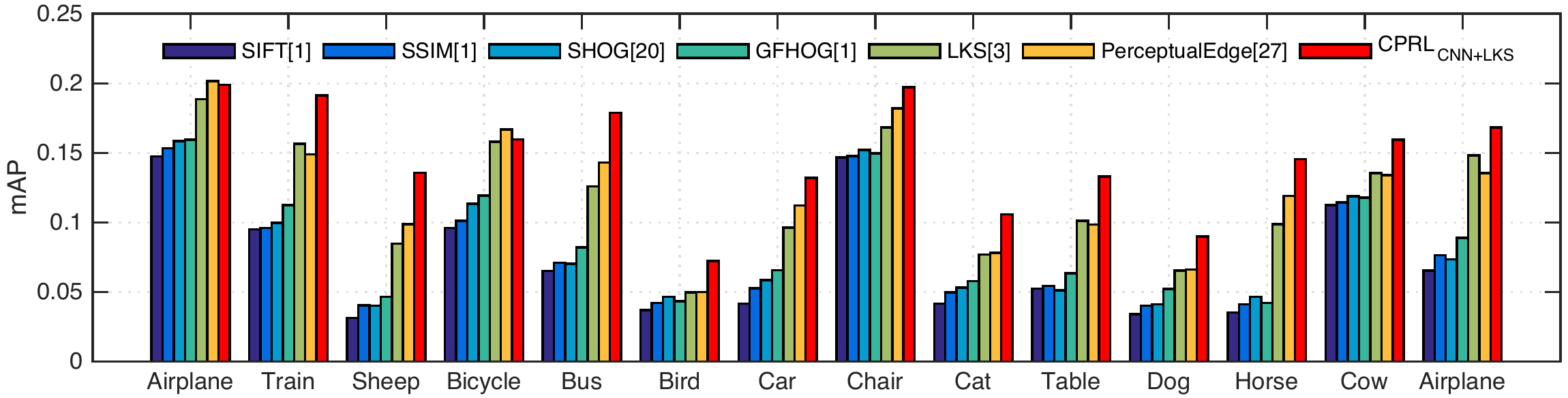} 
\caption{Retrieval performance comparison for each category of the Queen Mary SBIR dataset.}
\label{mapEachClass}
  \vspace{-13pt}
\end{figure*}

\begin{table}[t]
\centering
\caption{Comparison of different methods on Flickr15k and Queen Mary SBIR datasets\label{flickr15table}}
\resizebox{0.999\linewidth}{!} {
\begin{tabular}{lcc}
\toprule
\multirow{2}{*}{Method} & \multicolumn{2}{c}{mAP} \\\cmidrule{2-3}
                                      & Flick15k & QueenMary SBIR \\
\midrule
StructureTensor~\cite{eitz2010evaluation} & 0.0801 & 0.0601\\
SIFT~\cite{hu2013performance}  & 0.0967 & 0.0685\\
SSIM~\cite{hu2013performance} & 0.1068 &0.0745\\
SHOG~\cite{eitz2011sketch} &0.1152 & 0.0804\\
GFHOG~\cite{hu2013performance} & 0.1245 & 0.0858\\
LKS~\cite{saavedrasketch} & 0.1640 & 0.1182\\
PerceptualEdge~\cite{qi2015making} & 0.1741& 0.1246\\\midrule
SaN~\cite{yu2015sketch} & 0.1730 & 0.1211\\
Siamese CNN~\cite{qi2016sketch} & 0.1954 & - \\
\midrule \midrule
CPRL$_\textsc{GFHOG}$ with $f_{\textrm{SP}_B}$ & 0.1693  & 0.1103 \\
CPRL$_\textsc{LKS}$ with $f_{\textrm{SP}_B}$ ($\gamma=\mu=0$) &0.2278 & 0.1265\\
CPRL$_\textsc{LKS}$ with $f_{\textrm{SP}_B}$&0.2495 & 0.1467\\ 
CPRL$_\textsc{CNN+LKS}$ with $f_{\textrm{SP}_A}$ &\textbf{0.2659} & \textbf{0.1521}\\
CPRL$_\textsc{CNN+LKS}$ with $f_{\textrm{SP}_B}$ &\textbf{0.2734} & \textbf{0.1603}\\
\bottomrule
\end{tabular}
}
  \vspace{-13pt}
\end{table}

\begin{table}[t]
\centering
\caption{Comparison of different methods on the TU-Berlin Extension dataset.}
\resizebox{0.99\linewidth}{!} {
\begin{tabular}{lcc}
\toprule
Method & \tabincell{c}{Feature Dimension} & mAP \\
\midrule
SHOG~\cite{eitz2011sketch} & 1296 & 0.091 \\
GFHOG~\cite{hu2013performance} & 3500 & 0.119 \\
SHELO~\cite{saavedra2014sketch} & 1296 & 0.123 \\
LKS~\cite{saavedrasketch} & 1350 & 0.157 \\
\midrule
SaN~\cite{yu2015sketch} & 512 & 0.154 \\
Siamese CNN~\cite{qi2016sketch} & 64 & 0.322 \\
GN Triplet~\cite{sangkloy2016sketchy} & 1024 & 0.187\\
3D shape~\cite{wang2015sketch} & 64 & 0.054\\
DSH~\cite{liu2017deep} & 32 (bits) & \textbf{0.358}\\
DSH~\cite{liu2017deep} & 128 (bits) & \textbf{0.570}\\
\midrule
\midrule
CPRL$_\textsc{LKS}$ with $f_{\textrm{SP}_B}$ ($\gamma=\mu=0$) & 1000 & 0.269\\
CPRL$_\textsc{LKS}$ with $f_{\textrm{SP}_B}$& 1000 & 0.301\\ 
CPRL$_\textsc{CNN+LKS}$ with $f_{\textrm{SP}_A}$  & 1000 & 0.324\\ 
CPRL$_\textsc{CNN+LKS}$ with $f_{\textrm{SP}_B}$  & 1000 & 0.332\\
\bottomrule
\end{tabular}
}
\label{tuberlin}
  \vspace{-13pt}
\end{table}

\textbf{Results.} A performance comparison of different methods on the Flickr15k and the QueenMary SBIR datasets is shown in Table~\ref{flickr15table}, reporting the mean average 
precision (mAP), and in Figure~\ref{prcurve}, depicting the precision-recall (PR) curve. Analyzing results on the Flickr15K dataset, three observations
can be made:  (i) CPRL$_\textsc{CNN+LKS}$ achieves the best mAP, showing a significant performance improvement (9.93 points) compared to 
the best state of the art method ($0.1741$ of PerceptualEdge~\cite{qi2015making}); (ii) CPRL$_\textsc{GFHOG}$ and CPRL$_\textsc{LKS}$ compares favorably to GFHOG~\cite{hu2013performance} and 
LKS~\cite{saavedrasketch} with $4.48$ and $8.55$ points improvement respectively, meaning that the advantage of the academic learning paradigm and its 
instantiation under the framework of dictionary learning is clear and independent 
of the features used; (iii) The clear performance gap when CPRL$_\textsc{LKS}$ is compared 
to CPRL$_\textsc{LKS}$ ($\gamma=\mu=0$) demonstrates the effectiveness of the proposed CPPCL strategy.

The fact that the best performance is obtained with CPRL$_\textsc{CNN+LKS}$ confirms our original intuition that 
different features can represent better the two different modalities. 
Interestingly the results in the table also show that our approach outperforms the SaN method~\cite{yu2015sketch} and Siamese CNN method~\cite{qi2016sketch}, demonstrating the effectiveness of 
our framework in comparison with deep learning architectures.
Finally, Figure~\ref{flickqualitative} shows some qualitative 
results (top-five retrieved images) associated with the proposed method.

On the Queen Mary SBIR dataset, Table~\ref{flickr15table}, CPRL$_\textsc{CNN+LKS}$ achieves an mAP of 0.1603 which is 3.57 points better 
than the best of all the comparison methods. It should be noted that this is not a trivial improvement on this very challenging dataset. 
CPRL$_\textsc{CNN+LKS}$ also outperforms CPRL$_\textsc{LKS}$, demonstrating the effectiveness of using different 
descriptors for sketches and images in SBIR. We also believe that LKS features are not robust enough to represent objects
with various poses and cluttered background, as in Queen Mary SBIR dataset. CPRL$_\textsc{LKS}$ obtains a clear improvement 
over CPRL$_\textsc{LKS}$ ($\gamma=\mu=0$), 
further verifying the usefulness of the proposed CPPCL scheme. 
Additionally, we show the retrieval performance for the category-level retrieval task in Figure~\ref{mapEachClass}. It is clear 
that for most of the classes (except for Airplane and Bicycle), CPRL$_\textsc{CNN+LKS}$ significantly outperforms all the comparison methods. Finally, 
Figure~\ref{queenmaryexample} reports the top 5 retrieval results of CPRL$_\textsc{CNN+LKS}$ for 10 query samples of sketches.

We further verify our performance on a larger SBIR dataset TU-Berlin Extension. The results are shown in Table~\ref{tuberlin}. It is clear that CPRL$_\textsc{LKS}$ with $f_{\textrm{SP}_B}$ is significantly better than the LKS method and CPRL$_\textsc{KS}$ with $f_{\textrm{SP}_B}(\gamma=\mu=0)$, demonstrating the effectiveness of the proposed learning strategy. When using powerful CNNs features as input, our method obtains better performance than previous end-to-end trainable deep learning models \cite{yu2015sketch,qi2016sketch,sangkloy2016sketchy,wang2015sketch}. The DSH method in~\cite{liu2017deep} achieves the best performance by successfully combining deep networks with hashing. We believe that our learning strategy is complementary to their method and the idea of exploiting curriculum and self-paced learning in the context of deep hashing is an interesting direction for future works. 

\subsection{In-depth Analysis of CPRL}
\label{subsec:deep-anal}
In this section, we show the results of a further analysis of the proposed CPRL model on both the Flickr15k and the Queen Mary SBIR datasets. 
The analysis was conducted considering several aspects including sensitivity study, convergence analysis, effect of self-paced regularizers, 
impact of the curriculum construction and computational cost analysis.

\begin{figure*}[!t]
\centering
\includegraphics[width=0.99\textwidth]{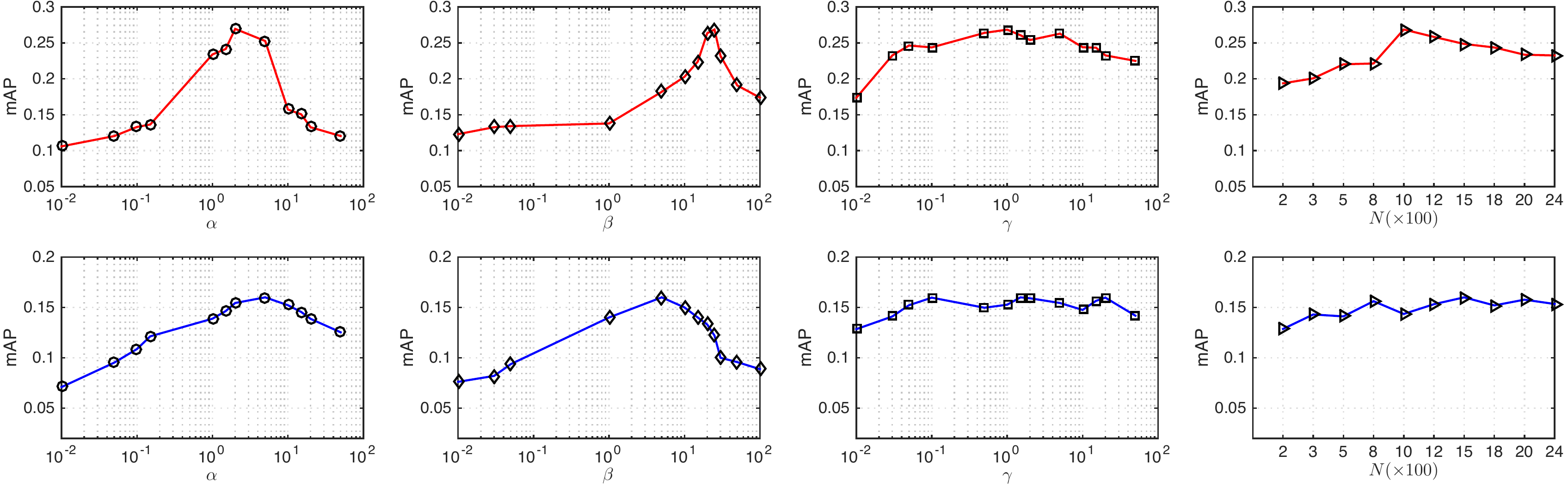} 
\caption{Empirical analysis of the model parameters: $\alpha$, $\beta$, $\gamma$ and the dictionary size $N$ on 
Flickr15k (first row) 
and Queen Mary SBIR (second row) datasets. 
}
\label{sensitivity}
  \vspace{-13pt}
\end{figure*}

\begin{figure}[t]
\centering
\includegraphics[width=0.4\textwidth]{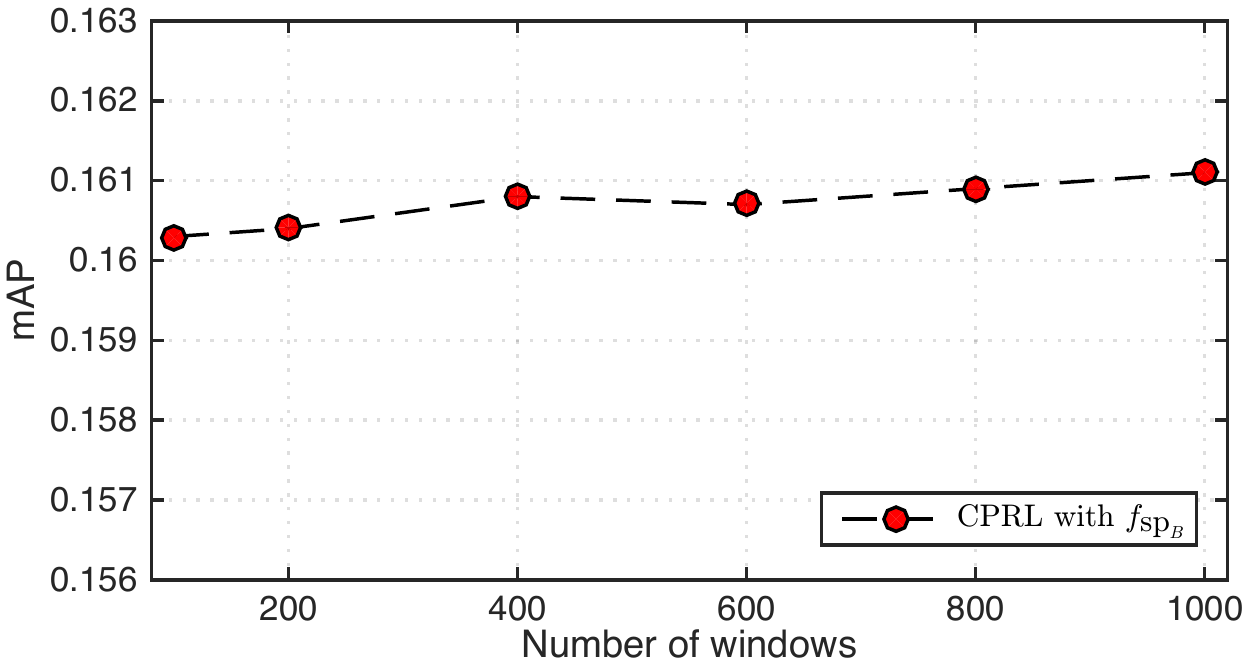} 
\caption{mAP at varying number of windows for edgeness calculation on the Queen Mary SBIR dataset.}
\label{windows}
  \vspace{-13pt}
\end{figure} 

\begin{figure}[!t]
\centering
\includegraphics[width =0.42\textwidth]{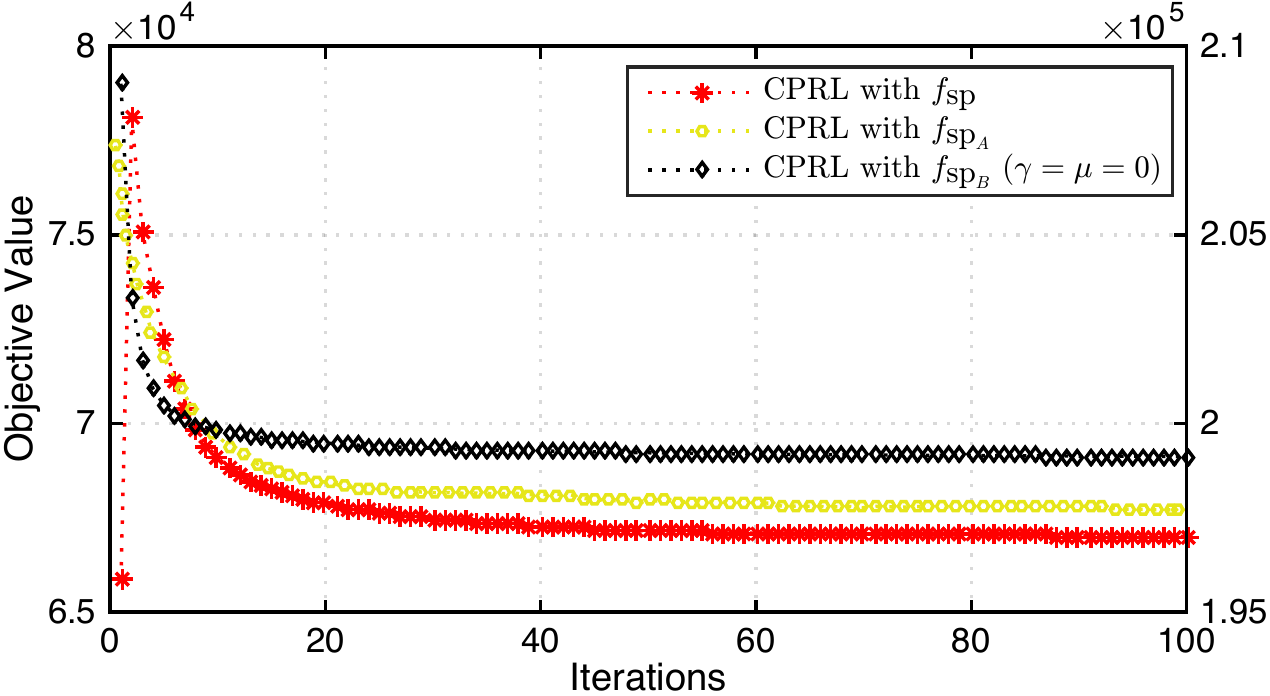}
\caption{Convergence in terms of objective value for: CPRL with $f_{\textsc{SP}_A}$ and with $f_{\textsc{SP}_B}$ and CPRL 
with $f_{\textsc{SP}_B}$ ($\gamma=\mu=0$).}
\label{fig:loss}
  \vspace{-13pt}
\end{figure}

\textbf{Sensitivity analysis.} 
First, we assess the influence in performance of the different model parameters in CPRL. Figure~\ref{sensitivity} shows the mAP 
as a function of the parameters $\alpha, \beta, \gamma, N$ on both the Flickr15k and the Queen Mary SBIR datasets.
The analysis on $\alpha, \beta$ and $\gamma$ is in the range $[10^{-2}, 10^2]$, on $N$ in 
the range $[200,2400]$. It is clear from the plots that, while the method is sensitive to $\alpha$ and $\beta$, its retrieval performance 
does not change drastically within a wide range of 
$\gamma$ and $N$. The sensitivity on $\beta$ was already observed in previous research works~\cite{huang2013coupled}. The 
performance trend varying the different parameters shows some similarity on both datasets. 
We also conduct an analysis to evaluate the impact on the performance of the number of windows used for the edgeness calculation 
in the sketch domain (Fig.~\ref{windows}). Fig.~\ref{windows} shows that the retrieval performance only slightly improves when increasing the number of windows. 
However, a large number of windows leads to a significant increase in terms of computational overhead. Therefore,
we set the number of windows equal to 100 in our experiments as it represents a good trade-off between accuracy and computational cost.

\textbf{Convergence analysis.}
Figure~\ref{fig:loss} plots the objective 
function value as a function of the iteration number for the proposed CPRL on Flickr15K with three different settings: 
(i) CPRL with $f_{\textrm{SP}_A}$; (ii) CPRL with $f_{\textrm{SP}_B}$ and (iii) CPRL with $f_{\textrm{SP}_B}$ ($\gamma=\mu=0$). The results 
clearly show the convergence of the proposed iterative optimization procedure. All the three settings of CPRL attain a stable solution 
within less than 40 iterations, proving the efficiency of the algorithm proposed to solve the CPRL optimization problem. It is worth noting 
that both CPRL with $f_{\textrm{SP}_A}$ and with $f_{\textrm{SP}_B}$ obtain a much lower local minima than 
CPRL ($\gamma=\mu=0$) (\eg~with $f_{\textrm{SP}_B}$ giving $6.8\times 10^4$ vs. $1.98\times 10^5$), verifying the beneficial
effect of the proposed CPPCL strategy for better optimization.

\begin{figure}[!t]
\centering
\includegraphics[width =0.42\textwidth]{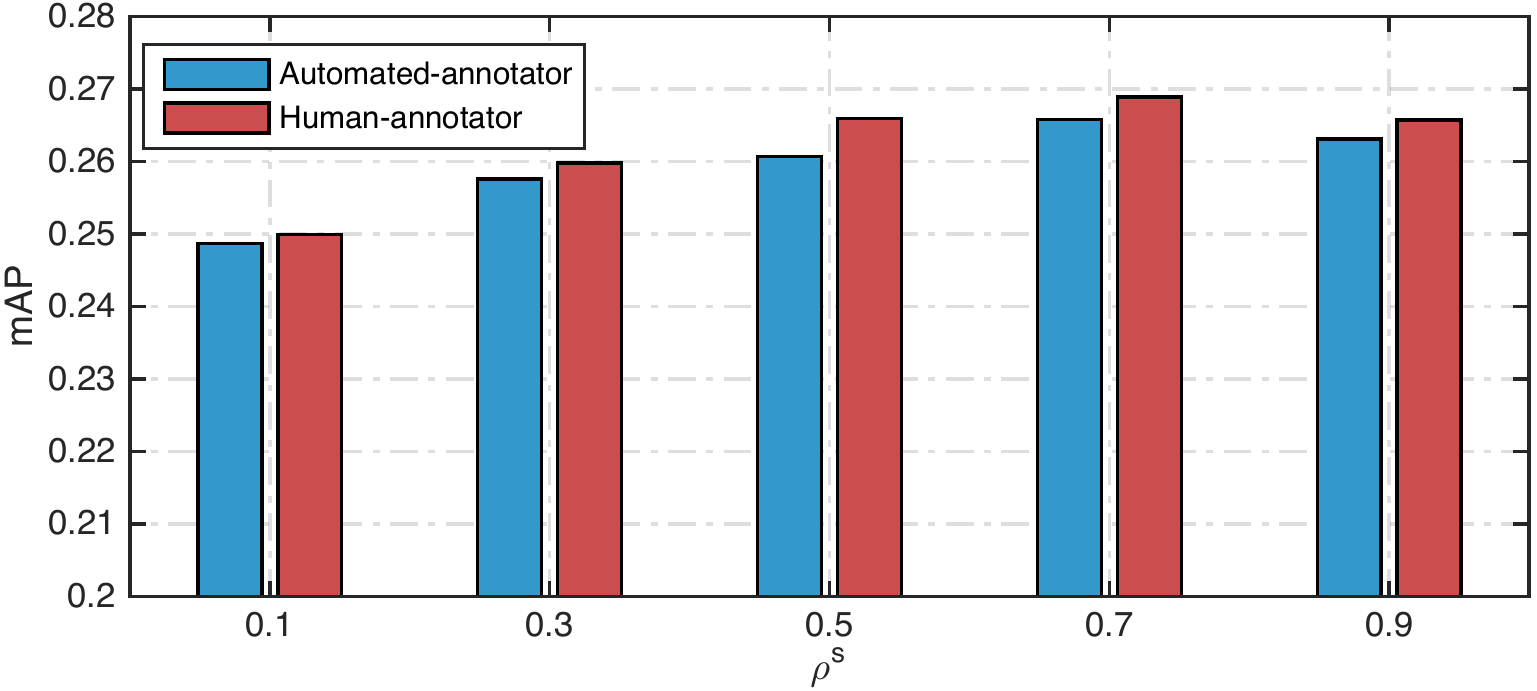}
\caption{Performance considering automatically obtained and human-annotated sketch partial curricula.}
\label{fig:curriculum_eva}
  \vspace{-13pt}
\end{figure}

\begin{figure}[!t]
\centering
\includegraphics[width =0.38\textwidth]{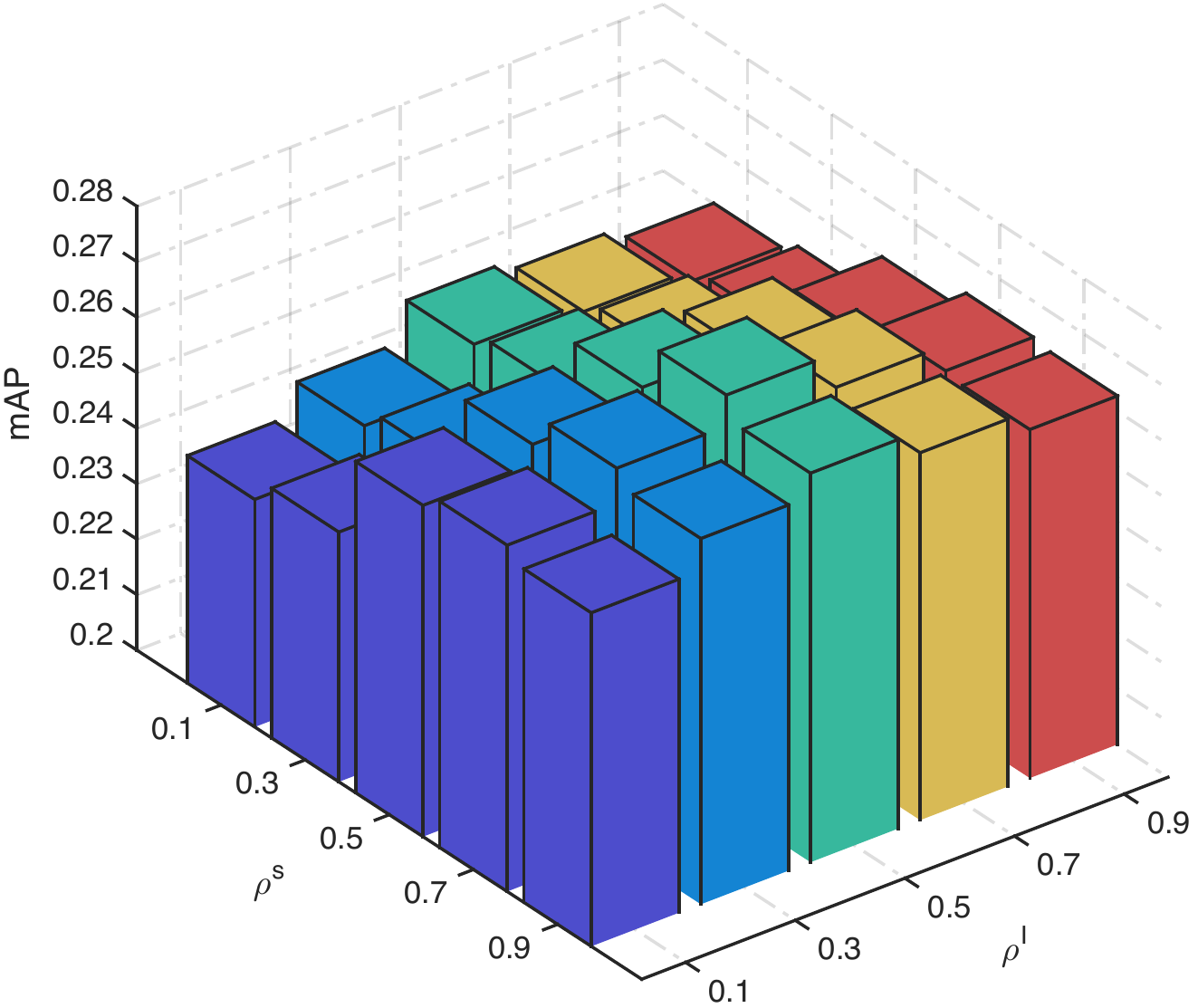}
\caption{Performance at varying the constraints ratio of the sketch and the image modalities.}
\label{fig:curriculum_3d}
  \vspace{-13pt}
\end{figure}

\textbf{Analysis of self-paced regularizers.} We carried out the retrieval experiments for CPRL with two different 
self-paced 
regularizers $f_{\textrm{SP}_A}$ and $f_{\textrm{SP}_B}$ on the four datasets. Table~\ref{sketch2photo} and Table~\ref{flickr15table} show 
the quantitative results of the two CPRL variants. We can observe that CPRL with $f_{\textrm{SP}_B}$ slightly outperforms 
CPRL with $f_{\textrm{SP}_A}$ on all the four datasets. We believe that this is probably due to the fact that when 
we optimize CPRL with $f_{\textrm{SP}_B}$, the self-pacing variables $\vect{V}$ are relaxed considering a 
real valued range [0, 1] (\ie~using a soft-weighting scheme) instead of discrete values. The soft weighting scheme is more effective 
than the hard weighting one in reflecting the true importance of samples in the training phase. This effect 
was previously observed in~\cite{zhao2015self, xumulti}. 

\textbf{Analysis of curriculum construction.} To investigate the influence the modalitiy-specific curricula to the final 
retrieval performance, we plot the mAP as a function of $\rho^\image$ and $\rho^\sketch$, the proportion of constraints used for the image 
and sketch curriculum relative to the number of possible constraints. Figure~\ref{fig:curriculum_3d} shows the plot with 
$\rho^\image$ and $\rho^\sketch$ taking five values ranging from 0.1 to 0.9. We can observe that for both modalities, the use of the 
curricula indeed helps boosting the performance, while using the excess of constraints leads to a slight decrease in performance.
This experimental finding supports our motivation of designing partial curricula learning in CPRL for SBIR. Our CPCL approach allows 
the human and the automated annotator to construct the partial curricula. To evaluate the difference of these two, 
Figure~\ref{fig:curriculum_eva} plots the mAP as a function of $\rho^\sketch$ with $\rho^\image$ fixed to be 0.3. It is 
clear that the human annotations correspond to more effective partial curricula, but yet the difference when compared with the 
automated curricula constructions is small.

\textbf{Computational cost analysis.} In the following, we analyze the computational time overhead on Flickr15K experiments both in the off-line 
training phase and during the online retrieval phase.
The training phase of our method mainly contains three steps: (i) feature extraction, (ii) curriculum construction and (iii) CPRL optimization. 
The input for CPRL are CNN features (for images) with size $4883 \times 2400$ and LKS features (for sketches) with size 
$132 \times 2400$, where $4833$ and $132$ are the number of training image and sketch samples respectively, and $2400$ is the feature dimension. 
Table~\ref{tab:time} reports computational times of different steps of the method. For the feature extraction, we consider 
CNN features from the image domain, which cost around $0.04$~seconds per image sample. The CNN image features were extracted with 
the GPU. LKS is used to extract features from the sketch domain. The automated curriculum construction takes 
around $8$~minutes and training CPRL and CPRL ($\gamma=\mu=0$) with $50$~iterations costs $27$~and $21$~minutes, respectively. 

\begin{figure}[t]
\centering
\includegraphics[width =0.42\textwidth]{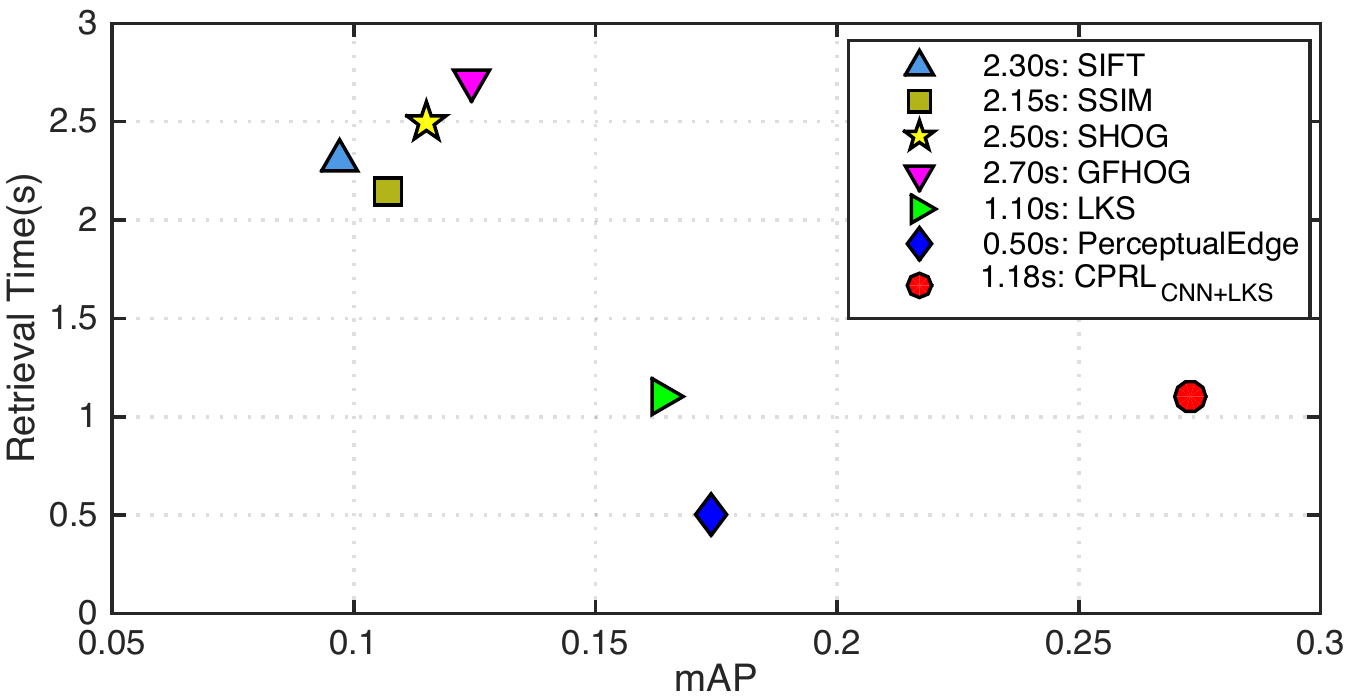}
\caption{Comparison of the average retrieval time of different methods with respect to the mAP.}
\label{fig:retrieval}
  \vspace{-13pt}
\end{figure}

\begin{table}[!t]
\caption{Computational cost of the different training steps. }
\centering
\resizebox{0.999\linewidth}{!} {
\begin{tabular}{lcc}
\toprule  
Phase &Component & Time overhead \\ 
\midrule
\multirow{2}{*}{\tabincell{c}{Feature Extraction}} 
&CNN (for images)   &  0.04 $\pm$ 0.01 sec/sample \\
&LKS  (for sketches)   &  1.1 $\pm$ 0.02 sec/sample\\
\midrule
\multirow{3}{*}{Training} & Curriculum Construction & 8 $\pm$ 1 min \\
&CPRL  &  27 $\pm$ 2 min \\
&CPRL($\gamma=\mu=0$) & 21$\pm$ 2 min \\
\midrule
Retrieval & CPRL & 1.1 $\pm$ 0.1 sec/sample \\ 
\bottomrule                      
\end{tabular}
}
\label{tab:time}
  \vspace{-13pt}
\end{table}

Online retrieval efficiency is a very important performance index for SBIR, especially for large-scale retrieval scenarios. 
Figure~\ref{fig:retrieval} plots the online retrieval time with respect to the mAP and compares CPRL with the state-of-the-art SBIR methods. 
Our CPRL$_\textsc{CNN+LKS}$ is based on three steps for the retrieval: (i) feature extraction from a query sketch sample using LKS, 
(ii) dictionary mapping to obtain a new feature representation and (iii) query the image features database with $k\textsc{-NN}$. The last two 
steps are very fast, and the feature extraction using LKS takes around 1 second. The average retrieval time for each query sample is around 
1.18 seconds. PerceptualEdge method achieves the best retrieval speed, as it uses only two steps namely the HOG feature extraction and 
direct matching. The retrieval speed of ours is comparable to the LKS method, and is almost 2 times faster than GFHOG, SHOG, SIFT and SSIM, which 
first extract features, and 
then construct bag-of-words descriptors and finally perform the retrieval. The reason is that the step of constructing the 
bag-of-words features is more time consuming than the dictionary mapping step. More importantly, our approach obtains a very good balance 
between the retrieval performance (mAP) and the computational efficiency.

To conclude, our approach achieves better or comparable speed than 
previous works based on direct feature matching. 
We believe that other strategies can be used to further speed up the retrieval process, such as adopting hash-based algorithms.
While this is not the focus of the current paper, our framework can be also extended in this direction.

\section{Conclusions}\label{sec:conclusion}
We presented a novel cross-domain representation learning framework for computing robust cross-modal
features for sketch-based image retrieval. In particular, this work explores
self-paced and curriculum learning schemes for dictionary learning. A novel cross-paced partial curriculum learning 
strategy is designed to learn from samples with an easy-to-hard order, such as to avoid bad local optimal into
dictionary learning optimization.
The proposed framework naturally handles different descriptors for the sketch and the image domains. Therefore, domain-specific discriminative
feature representations (\eg, CNN features for images) are considered, overcoming the limitations of previous works. Extensive evaluation on four publicly available datasets shows that our approach achieves very competitive performance over state-of-the art methods for SBIR. 

In this paper CPPCL is instantiated within a coupled dictionary learning model for addressing the SBIR task. However, CPPCL 
is a general strategy which can be also combined with other 
representation learning methods. Future works will explore the adoption of CPPCL into deep cross-domain~\cite{xu2015learning} and deep structured learning models~\cite{xu2017learningnips}.

\ifCLASSOPTIONcaptionsoff
  \newpage
\fi

\bibliographystyle{IEEEtran}
\bibliography{egbib}

\begin{IEEEbiography}
{Dan Xu} is a Ph.D. candidate in the Department of Information Engineering and Computer Science, and a member of Multimedia and Human Understanding Group (MHUG) led by Prof. Nicu Sebe at the University of Trento. He was a research assistant in the Multimedia Laboratory in the Department of Electronic Engineering at the Chinese University of Hong Kong. 
His research focuses on computer vision, multimedia and machine learning. Specifically, he is interested in deep learning, structured prediction and cross-modal representation learning and the applications to scene understanding tasks. He received the Intel best scientific paper award at ICPR 2016.
\end{IEEEbiography}
\vspace{-1.3cm}

\begin{IEEEbiography}
{Xavier Alameda-Pineda} received the M.Sc. degree in mathematics and telecommunications engineering from the Universitat Politecnica de Catalunya BarcelonaTech in 2008 and 2009 respectively, the
M.Sc. degree in computer science from the Universite Joseph Fourier and Grenoble INP in 2010, and the Ph.D. degree in mathematics/computer science from the Universite Joseph Fourier in 2013. He worked towards his Ph.D. degree in the Perception Team, at INRIA Grenoble Rhone-Alpes. He currently holds a postdoctoral position at the Multimodal Human Understanding Group at University of Trento. His research interests are machine learning and signal processing for scene understanding, speaker diaritzation and tracking, sound source separation and behavior analysis.
\end{IEEEbiography}
\vspace{-1.3cm}

\begin{IEEEbiography}
{Jingkuan Song} received the B.S. degree in computer
science from the University of Electronic Science
and Technology of China and the Ph.D. degree
in information technology from The University of
Queensland, Australia, in 2014. He is currently
a Post-Doctoral Research Scientist with Columbia
University. He joined the University of Trento as a
Research Fellow sponsored by Prof. Nicu Sebe from
2014-2016. His research interest includes large-scale
multimedia retrieval, image/video segmentation,
and image/video annotation using hashing,
graph learning, and deep learning techniques.
\end{IEEEbiography}
\vspace{-1.3cm}

\begin{IEEEbiography}
{Elisa Ricci}
received the PhD degree from the
University of Perugia in 2008. She is an assistant
professor at the University of Perugia and a
researcher at Fondazione Bruno Kessler. She
has since been a post-doctoral researcher at
Idiap, Martigny, and Fondazione Bruno Kessler,
Trento. She was also a visiting researcher at the
University of Bristol. Her research interests are
mainly in the areas of computer vision and
machine learning. She is a member of the IEEE.
\end{IEEEbiography}
\vspace{-1.3cm}

\begin{IEEEbiography}
{Nicu Sebe} is Professor with the University of
Trento, Italy, leading the research in the areas of
multimedia information retrieval and human behavior
understanding. He was the General Co-
Chair of the IEEE FG Conference 2008 and ACM
Multimedia 2013, and the Program Chair of the
International Conference on Image and Video
Retrieval in 2007 and 2010, ACM Multimedia
2007 and 2011. He is the Program Chair of ICCV
2017 and ECCV 2016, and a General Chair of
ACM ICMR 2017 and ICPR 2020. He is a fellow
of the International Association for Pattern Recognition.
\end{IEEEbiography}
%






\end{document}